\documentclass[10pt,twocolumn,letterpaper]{article}

\usepackage{iccv}
\usepackage{times}

\iccvfinalcopy 

\def\iccvPaperID{****} 

\usepackage{graphicx}
\usepackage{amsmath}
\usepackage{amssymb}
\usepackage{booktabs}
\usepackage{subfig}
\usepackage[font=small]{caption}

\usepackage{diagbox}
\usepackage{nicefrac}       %
\usepackage{microtype}      %

\usepackage{blindtext}
\usepackage{float}
\usepackage{enumitem}
\usepackage[table]{xcolor}
\usepackage{tabulary,multirow,overpic}
\usepackage{verbatim}
\usepackage{color}
\usepackage{multicol}
\usepackage{dblfloatfix}
\usepackage{pifont}%
\usepackage[accsupp]{axessibility}  %
\usepackage{makecell}

\usepackage{marvosym}

\usepackage{multicol}

\usepackage{fontawesome5}

\newlength\savewidth\newcommand\shline{\noalign{\global\savewidth\arrayrulewidth\global\arrayrulewidth 1pt}\hline\noalign{\global\arrayrulewidth\savewidth}}
\newcommand{\tablestyle}[2]{\setlength{\tabcolsep}{#1}\renewcommand{\arraystretch}{#2}\centering\footnotesize}

\newcolumntype{x}[1]{>{\centering\arraybackslash}p{#1pt}}
\newcolumntype{y}[1]{>{\raggedright\arraybackslash}p{#1pt}}
\newcolumntype{z}[1]{>{\raggedleft\arraybackslash}p{#1pt}}

\definecolor{baselinecolor}{gray}{.92}

\definecolor{demphcolor}{gray}{.75}
\newcommand{\demph}[1]{\textcolor{demphcolor}{#1}}

\definecolor{demphcolorinline}{gray}{.3}
\newcommand{\demphinline}[1]{\textcolor{demphcolorinline}{#1}}

\definecolor{demphcolor1}{gray}{.6}
\newcommand{\demphs}[1]{\textcolor{demphcolor1}{#1}}

\definecolor{eva01purple}{RGB}{168,119,200}
\newcommand{\evapurple}[1]{\textcolor{eva01purple}{#1}}

\definecolor{eva01green}{RGB}{82,208,83}

\definecolor{eva02red}{RGB}{236,35,35}
\newcommand{\evared}[1]{\textcolor{eva02red}{#1}}

\definecolor{eva02yellow}{RGB}{249,157,83}

\newcommand{\demphsye}[1]{\textcolor{eva02yellow}{#1}}

\definecolor{02pink}{RGB}{240,178,188}

\newcommand{\ph}[1]{\textcolor{white}{#1}}
\newcommand{\phpink}[1]{\textcolor{02pink!20}{#1}}

\definecolor{citecolor}{RGB}{34,139,34}
\definecolor{citecolor2}{HTML}{0071bc}
\definecolor{Graylight}{gray}{0.9}
\definecolor{lightred}{RGB}{241,140,142}

\usepackage[pagebackref=true,breaklinks=true,colorlinks,citecolor=eva02yellow,urlcolor=eva02red,bookmarks=false]{hyperref}

\definecolor{clipbaselinecolor}{gray}{.9}

\definecolor{defaultcolor}{HTML}{E8E2F7}
\newcommand{\evadefault}[1]{\cellcolor{02pink!20}{#1}}

\renewcommand{\paragraph}[1]{\vspace{1.25mm}\noindent\textbf{#1}}

\newcommand{\app}{\raise.17ex\hbox{$\scriptstyle\sim$}}
\newcommand{\appp}{\raise.20ex\hbox{$\scriptscriptstyle\sim$}}

\def\x{\raise.15ex\hbox{$\times$}}
\newcommand{\upp}{\raise.50ex\hbox{$\scriptscriptstyle\nearrow$}}

\newcommand{\figref}[1]{Fig.~\ref{#1}}
\newcommand{\tblref}[1]{Table~\ref{#1}}
\newcommand{\sref}[1]{\S\ref{#1}}

\newcommand{\boxAP}{AP$^\text{box}$\xspace}
\newcommand{\maskAP}{AP$^\text{mask}$\xspace}

\newcommand{\val}{$\mathtt{val}$\xspace}
\newcommand{\test}{$\mathtt{test}$-$\mathtt{dev}$\xspace}
\newcommand{\tset}{$\mathtt{test}$\xspace}

\newcommand{\mIoU}{mIoU\xspace}
\newcommand{\mIoUss}{mIoU$^\text{ss}$\xspace}
\newcommand{\mIoUms}{mIoU$^\text{ms}$\xspace}

\newcommand{\evaone}{{\textbf{\evapurple{EVA}}}\xspace}
\newcommand{\eva}{{\textbf{\evared{EVA-02}}}\xspace}
\newcommand{\trv}{{\textbf{\evared{TrV}}}\xspace}

\newcommand{\rpink}{\rowcolor{02pink!20}}

\newcommand{\suptext}[1]{$^{\text{#1}}$}

\newcommand{\cmark}{\ding{51}}%
\newcommand{\xmark}{\ding{55}}%

\def \alambic {\includegraphics[width=0.02\linewidth]{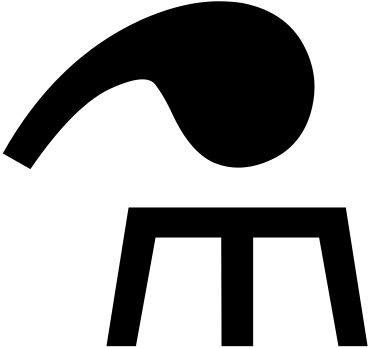}\xspace}

\usepackage[capitalize]{cleveref}
\crefname{section}{Sec.}{Secs.}
\Crefname{section}{Section}{Sections}
\Crefname{table}{Table}{Tables}
\crefname{table}{Tab.}{Tabs.}

\begin{document}

\title{\eva: A Visual Representation for Neon Genesis}

\newcommand{\authorskip}{\hspace{4mm}}

\author{
{
Yuxin Fang\textsuperscript{2,1} \authorskip
Quan Sun\textsuperscript{1} \authorskip 
Xinggang Wang\textsuperscript{2} \authorskip 
Tiejun Huang\textsuperscript{1} \authorskip 
Xinlong Wang\textsuperscript{1} \authorskip 
Yue Cao\textsuperscript{1}
}
\\[0.5mm]
{
\fontsize{10.0pt}{9.84pt}\selectfont
\textsuperscript{1}Beijing Academy of Artificial Intelligence \hspace{5.5mm} \textsuperscript{2}Huazhong University of Science and Technology 
}
\\[3.0mm]
{
\fontsize{9.pt}{9.84pt}\selectfont
\textbf{Fight together with \href{https://en.wikipedia.org/wiki/Asuka_Langley_Soryu}{\color{eva02red}Asuka} at }\href{https://github.com/baaivision/EVA/tree/master/EVA-02}{\color{eva02red} \bfseries \ttfamily baaivision/EVA/02}
}
}

\maketitle

\begin{abstract}
   We launch \eva, a next-generation Transformer-based visual representation pre-trained to reconstruct strong and robust language-aligned vision features via masked image modeling. With an updated plain Transformer architecture as well as extensive pre-training from an open \& accessible giant CLIP vision encoder, \eva demonstrates superior performance compared to prior state-of-the-art approaches across various representative vision tasks, while utilizing significantly fewer parameters and compute budgets. Notably, using exclusively publicly accessible training data, \eva with only \textbf{304M} parameters achieves a phenomenal \textbf{90.0} fine-tuning top-1 accuracy on ImageNet-1K val set. Additionally, our \eva-CLIP can reach up to \textbf{80.4} zero-shot top-1 on ImageNet-1K, outperforming the previous largest \& best open-sourced CLIP with only \app1/6 parameters and \app1/6 image-text training data. We offer four \eva variants in various model sizes, ranging from 6M to 304M parameters, all with impressive performance. To facilitate open access and open research, we release the complete suite of \eva to the community.
\end{abstract}


\begin{table*}[!t] 
\vspace{-2.5em}
    \centering
    \tablestyle{2pt}{1.2}
    \begin{tabular}{c|r|x{35}x{35}|x{55}|x{25}x{55}|x{30}x{30}|x{30}x{30}|x{30}x{30}}
        & & \multicolumn{3}{c|}{\textbf{zero-shot evaluation} with {\scriptsize{EVA-CLIP}}} & \multicolumn{8}{c}{\textbf{transfer learning}} \\
        & & \multicolumn{2}{c|}{image cls} &  video cls & \multicolumn{2}{c|}{e2e ft image cls} & \multicolumn{2}{c|}{object det} & \multicolumn{2}{c|}{instance seg} & \multicolumn{2}{c}{semantic seg}\\
        & enc. & \scriptsize IN-1K & \scriptsize 27 avg. & \scriptsize 4 avg. & \scriptsize IN-1K & \scriptsize variants avg. & \scriptsize COCO & \scriptsize LVIS & \scriptsize COCO & \scriptsize LVIS & \scriptsize COCO164K & \scriptsize ADE20K \\
        method & \#params & \scriptsize (\tblref{tab: clip configs and 1k zs}) & \scriptsize (\tblref{tab: clip zs img cls 27}) & \scriptsize (\tblref{tab: clip zs video cls}) & \scriptsize (\tblref{tab: 1k cls}) & \scriptsize (\tblref{tab: cls rob and gen}) & \scriptsize (\tblref{tab: large od}) & \scriptsize (\tblref{tab: large od}) & \scriptsize (\tblref{tab: large od}) & \scriptsize (\tblref{tab: large od}) & \scriptsize (\tblref{tab: m2f sem seg}) & \scriptsize (\tblref{tab: m2f sem seg}) \\
        \shline
        \scriptsize \evaone~\cite{eva} & 1011M & 78.5 & 71.4 & 66.0 & 89.7 & 84.0 & 64.4 & 62.2 & 55.5 & 55.0 & 53.4 & \textbf{62.3} \\
        \rpink
        \scriptsize \eva-L & 304M & \textbf{80.4} & \textbf{73.5} & \textbf{67.7} & \textbf{90.0} & \textbf{85.2} & \textbf{64.5} & \textbf{65.2} & \textbf{55.8} & \textbf{57.3} & \textbf{53.7} & 62.0 \\
        \demphsye{$\Delta$} & \bf \scriptsize \demphsye{-707M} & \bf \scriptsize \demphsye{+1.9} & \bf \scriptsize \demphsye{+2.1} & \bf \scriptsize \demphsye{+1.7} & \bf \scriptsize \demphsye{+0.3} & \bf \scriptsize \demphsye{+1.2} & \bf \scriptsize \demphsye{+0.1} & \bf \scriptsize \demphsye{+3.0} & \bf \scriptsize \demphsye{+0.3} & \bf \scriptsize \demphsye{+2.3} & \bf \scriptsize \demphsye{+0.3} & \scriptsize \demphs{-0.3}        
        \end{tabular}
        \vspace{-1.em}
        \caption{\textbf{Quantitative summary of \eva-L's performance on various mainstream vision benchmarks.}}
        \vspace{-1.em}
        \label{tab: summary of performance}
\end{table*}

\section{Introduction}
\label{sec: intro}

Recent research advancements have led to a surge of interest in scaling up vision~\cite{swinv2,eva,internimage,revcol} as well as vision-language~\cite{yuan2021florence,beit3,cliph,yu2022coca} representations.
These efforts are driven by the belief that increasing the number of parameters, data, and compute budgets will ultimately result in improved performance~\cite{kaplan2020scaling,zhai2022scalingvit,xie2022datascaling,basic}.

However, there is an increasing gap in computer vision between large-scale models that achieve state-of-the-art performance and models that are affordable for the wider research community.
Training, tuning, and evaluating very large vision models requires significant computational resources, which can be prohibitively expensive and time-consuming.
This usually leads to large-scale visual representations being trained in a few-shot or even single-shot manner, limiting the ability to fully optimize the entire process.
In addition, the study of state-of-the-art representations is frequently conducted using huge amounts of infrastructure and web-scale private training data~\cite{zhai2022scalingvit,alayrac2022flamingo,chen2022pali,vit22b}, which makes it difficult to evaluate the effects of modeling advancements in a feasible and transparent way, and restricts access to a broad range of researchers and practitioners.
These challenges highlight a pressing need for a more efficient and accessible approach of training and evaluating state-of-the-art vision as well as vision-language representations.

\begin{figure}
    \centering
    \includegraphics[width=\linewidth]{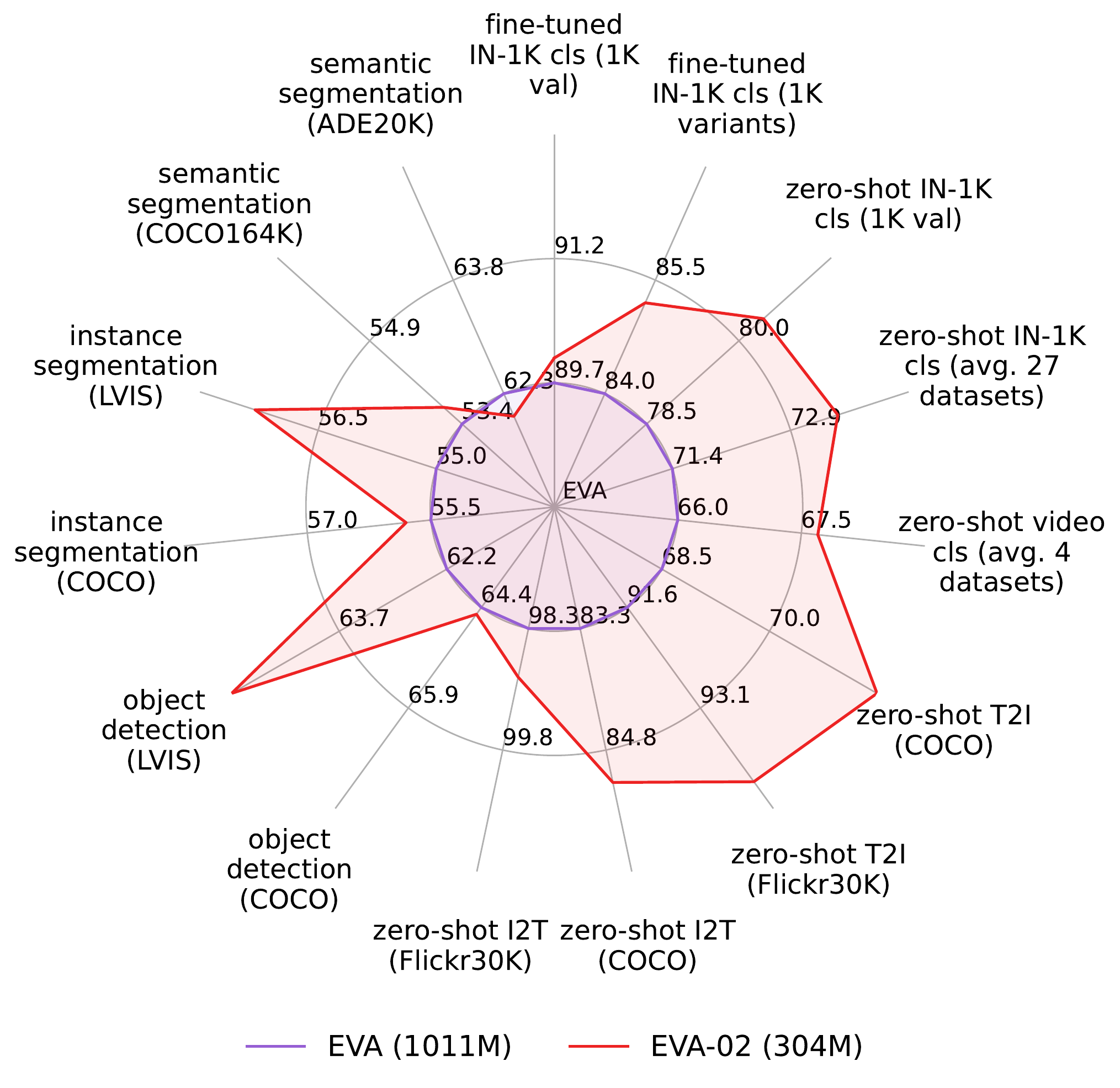}
    \vspace{-2.5em}
    \caption{Qualitative comparisons between \eva (\#params: 304M) and \evaone (\#params: 1011M)~\cite{eva} pre-trained representations. \eva with only 304M pre-trained representations pulls off a ``giant-killing'' act against the previous state-of-the-art \evaone. \\ {\footnotesize{\demphinline{ $*$ Notice that the scale of each axis in the radar chart is normalized by the performance of \evaone, and the stride of each axis are the same}}}}
    \label{fig: teaser}
    \vspace{-.6em}
\end{figure}

In this work, we present \eva, a series of robustly optimized plain Vision Transformers (ViTs)~\cite{vaswani2017attention,dosovitskiy2020vit} with moderate model sizes that are equipped with transferable bidirectional visual representations~\cite{devlin2018bert,liu2019roberta} learned from a strong  CLIP~\cite{clip,eva} vision encoder via masked image modeling (MIM) pre-training~\cite{bao2021beit}.
Compared with current leading vision models with billions of parameters~\cite{swinv2,eva,internimage,revcol}, these \eva variants require far fewer compute budgets and resources to investigate, allowing for an in-depth exploration of often-overlooked aspects.

Our empirical investigation indicates that the smaller-sized plain ViTs are highly capable, and their potential has been significantly underestimated.
By leveraging the latest plain Transformer architecture design~\cite{dauphin2017language,shazeer2020glu,su2021roformer,wang2022foundation} borrowed from language models, as well as thorough MIM pre-training from a publicly available giant \evaone-CLIP~\cite{eva} vision encoder, \eva is able to achieve superior performance compared to prior state-of-the-art approaches with much larger model sizes on various visual tasks.

\begin{table}[!b]
    \vspace{-1.5em}
    \centering
    \tablestyle{3.8pt}{1.2}
    \begin{tabular}{c|x{30}x{30}x{30}x{40}|r}
        & & & & & {\scriptsize{IN-1K}} ft \ph{.} \\
        arch. & norm & init. & \scriptsize FFN & pos. embed. & top-1 acc. \\
        \shline
        \multicolumn{6}{c}{\scriptsize{{base}-sized model (86M), IN-1K ft number of tokens = 196}} \\
        \hline
        & \scriptsize \demph{pre-LN} & \scriptsize \demph{BEiT} & \scriptsize \demph{MLP} & \demph{abs. {\scriptsize{PE}}} & \scriptsize 84.0 ($*$) \\
        & \scriptsize \demph{pre-LN} & \scriptsize xnorm & \scriptsize \demph{MLP} & \demph{abs. {\scriptsize{PE}}} & \scriptsize 84.0 \ph{($*$)} \\
        & \scriptsize \demph{pre-LN} & \scriptsize \demph{BEiT} & \scriptsize SwiGLU & \demph{abs. {\scriptsize{PE}}} & \scriptsize 83.9 \ph{($*$)} \\
        & \scriptsize \demph{pre-LN} & \scriptsize xnorm & \scriptsize SwiGLU & \demph{abs. {\scriptsize{PE}}} & \scriptsize 85.0 \ph{($*$)} \\
        & \scriptsize sub-LN & \scriptsize xnorm & \scriptsize SwiGLU & \demph{abs. {\scriptsize{PE}}} & \scriptsize 85.2 \ph{($*$)} \\
        \rpink
        \scriptsize \trv & \scriptsize sub-LN & \scriptsize xnorm & \scriptsize SwiGLU & \scriptsize 2D RoPE & \scriptsize \textbf{85.6} ($\dag$) \\
        & \scriptsize {sub-LN} & \scriptsize {xnorm} & \scriptsize {SwiGLU} & \demph{{\scriptsize{2D}} rel. {\scriptsize{PE}}} & \scriptsize {\xmark} \ph{($*$)} \\
        & \scriptsize \demph{post-LN} & \scriptsize {xnorm} & \scriptsize {SwiGLU} & \scriptsize RoPE & \scriptsize {\xmark} \ph{($*$)} \\
    \end{tabular}
    \vspace{-1em}
    \caption{\textbf{From ViT to \trv.} All experiments are conducted with the base-sized plain ViT {\footnotesize \demphinline{(macro architecture: depth=12, width=768, \#heads=12)}} with 300-epoch MIM pre-training on IN-1K. The MIM objective is to reconstruct the masked-out \evaone-CLIP vision features based on visible image patches. ``\xmark'': unstable or diverged pre-training. ``xnorm'': $\mathtt{xavier}$ $\mathtt{normal}$ weight initialization.}
    \vspace{-2.5em}
    \label{tab: from vit to trv}
\end{table}

Remarkably, using exclusively 38 million publicly accessible data, the small-sized variant of \eva with only 22M parameters achieves 85.8 fine-tuning top-1 accuracy on ImageNet-1K (IN-1K) val set~\cite{russakovsky2015imagenet}, while the large model with only \textbf{304M} parameters achieves an outstanding \textbf{90.0} fine-tuning top-1 accuracy.
Moreover, we show that initializing the image encoder of a CLIP via MIM pre-trained \eva representations can reach up to \textbf{80.4} zero-shot top-1 on IN-1K val, outperforming the previous largest \& best open-sourced CLIP-Giant~\cite{clipbigg} with only \app1/6 parameters and \app1/6 image-text training data.
\eva also achieves state-of-the-art performances on other representative vision tasks such as object detection and instance segmentation on LVIS~\cite{gupta2019lvis} {\footnotesize{\demphinline{(65.2 \boxAP \& 57.3 \maskAP on \val)}}} and COCO~\cite{lin2014coco} {\footnotesize{\demphinline{(64.5 \boxAP \& 55.8 \maskAP on \test)}}}, as well as semantic segmentation on COCO-stuff-164K~\cite{coco-stuff} {\footnotesize{\demphinline{(53.7 \mIoUss)}}} and ADE20K~\cite{zhou2018ade} {\footnotesize{\demphinline{(61.7 \mIoUss and 62.0 \mIoUms)}}}. 
For a quantitative summary of \eva's performance, please refer to \tblref{tab: summary of performance}.

The proposed \eva series offers a diverse range of model sizes, ranging from 6M to 304M parameters, each demonstrating exceptional performance.
The aim of this work is not necessarily to propose a novel method, but strive to identify a robust and effective recipe for making state-of-the-art models more affordable in practice.
By providing a more accessible and performant option, \eva democratizes access to state-of-the-art vision models, allowing researchers as well as practitioners to conduct high-quality research without the need for extensive infrastructure or resources.
We hope our efforts enable a broader range of the research community to advance the field in a more efficient and equitable manner.

\section{Approach}
\label{sec: approach}

The aim of \eva is to introduce a next-generation Transformer-based visual representation that achieves strong performances with moderate model sizes. 
To achieve this goal, our representation instrumentality project consists of two parts: architectural improvements made to the plain ViT in~\sref{sec: architecture}, as well as our MIM pre-training strategy in~\sref{sec: pre-training strategy}.

\begin{figure}[!b]
    \vspace{-1.5em}
    \centering
    \includegraphics[width=\linewidth]{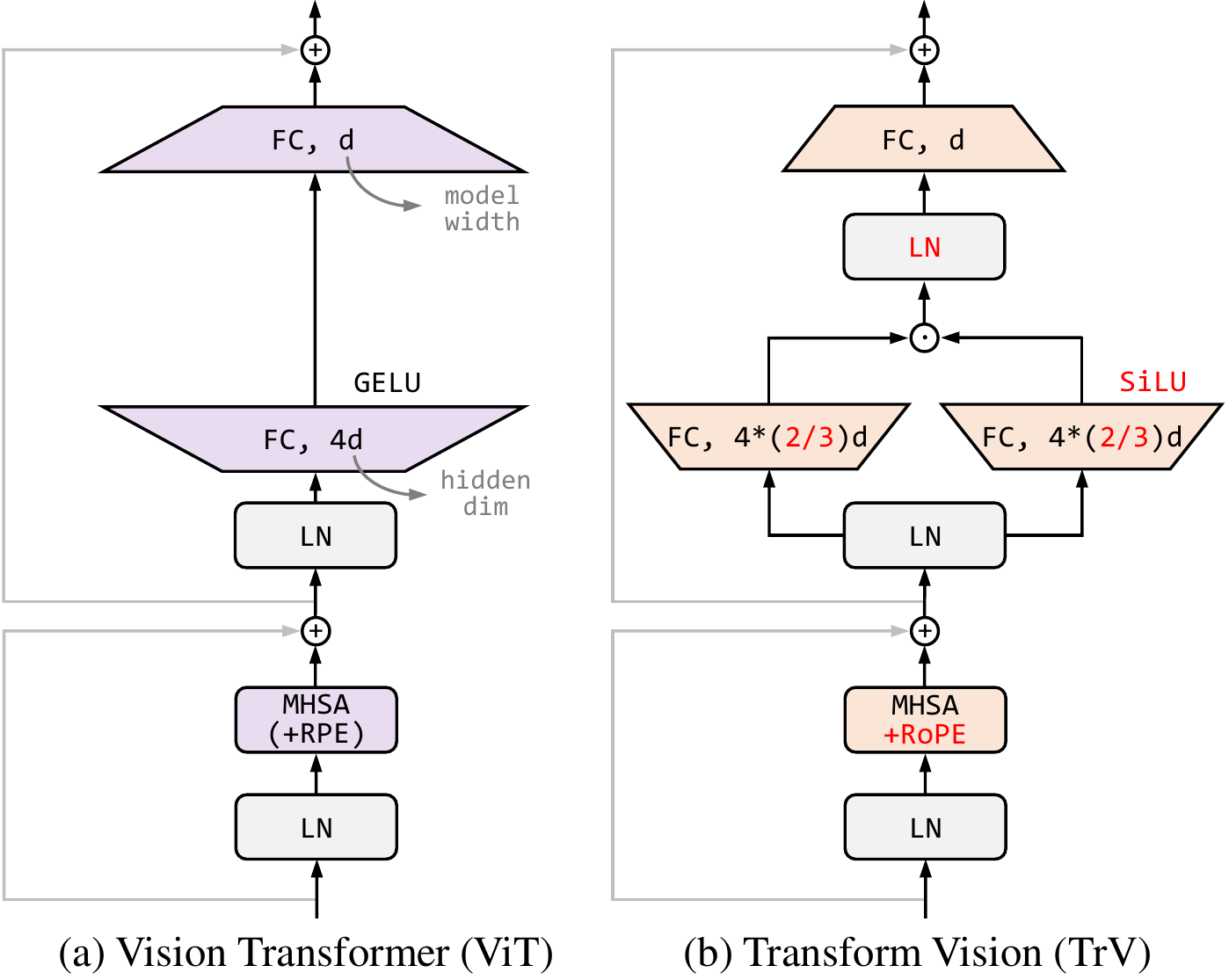}
    \vspace{-2em}
    \caption{\textbf{An illustration of ViT and \trv blocks.} \trv builds upon the original plain ViT architecture~\cite{dosovitskiy2020vit} and includes several enhancements: SwiGLU FFN, sub-LN, 2D RoPE, and $\mathtt{xavier}$ $\mathtt{normal}$ weight initialization. To keep the parameter \& FLOPs consistent with the baseline, the FFN hidden dim of SwiGLU is 2/3$\times$ of the typical MLP counterpart.}
    \label{fig: vit and trv}
    \vspace{-2.em}
\end{figure}

\subsection{Architecture}
\label{sec: architecture}

At a high level, \textit{plain} ViT along with its variants comes with interleaved multi-head self-attention (MHSA) layers for global spatial information aggregation \& position-wise feedforward networks (FFNs) for feature transformation, \textit{without} downsampling layers and multi-stage design~\cite{vaswani2017attention,dosovitskiy2020vit,touvron2021training}.
This makes it an ideal testbed for representation learning due to its minimal visual structure prior and biases, as well as its natural compatibility with masked modeling, which is proven to be a simple, strong, and scalable pre-training approach~\cite{bao2021beit,beitv2,beit3,eva}.
Pre-trained plain ViT can also be successfully adapted to challenging vision tasks that require high-resolution inputs \& multi-scale representations with feasible costs~\cite{li2022exploring,fang2022unleashing}.

\begin{table}[!t]
\vspace{-2.5em}
    \centering
    \tablestyle{1.8pt}{1.2}
    \begin{tabular}{c|c|crc|r}
        & \scriptsize MIM & & & \scriptsize IN-21K & {\scriptsize{IN-1K}} ft \ph{.} \\
        arch. & teacher & pt dataset & pt epochs \ph{+} & intermed. ft & top-1 acc. \\
        \shline
        \multicolumn{6}{c}{\scriptsize{(a) base-sized model (86M), IN-1K ft number of tokens = 196}} \\
        \hline
        \scriptsize ViT-B & \scriptsize VQKD-B~\cite{beitv2} & \scriptsize IN-1K & \scriptsize 300 (0.2M-step) & \scriptsize \xmark & \scriptsize 85.0 \ph{($*$)} \\
        \scriptsize ViT-B & \scriptsize CLIP-B~\cite{clip} & \scriptsize IN-1K & \scriptsize 300 (0.2M-step) & \scriptsize \xmark & \scriptsize 85.0 \ph{($*$)} \\
        \scriptsize ViT-B & \scriptsize \evaone-CLIP~\cite{eva} & \scriptsize IN-1K & \scriptsize 300 (0.2M-step) & \scriptsize \xmark & \scriptsize 84.0 ($*$) \\
        \rpink
        \scriptsize \trv-B & \scriptsize \evaone-CLIP~\cite{eva} & \scriptsize IN-1K & \scriptsize 300 (0.2M-step) & \scriptsize \xmark & \scriptsize 85.6 ($\dag$) \\
        \shline
        \multicolumn{6}{c}{\scriptsize{(b) base-sized model, \textbf{longer} pre-training}} \\
        \hline
        \scriptsize ViT-B & \scriptsize VQKD-B~\cite{beitv2} & \scriptsize IN-1K & \scriptsize 1600 (1M-step) & \scriptsize \xmark & \scriptsize 85.5 \ph{($*$)} \\
        \rpink
        \scriptsize \trv-B & \scriptsize \evaone-CLIP~\cite{eva} & \scriptsize IN-1K & \scriptsize 1600 (1M-step) & \scriptsize \xmark & \scriptsize 86.8 \phpink{($*$)} \\
        \shline
        \multicolumn{6}{c}{\scriptsize{(c) base-sized model, \textbf{longer} pre-training \& \textbf{larger} dataset}} \\
        \hline
        \scriptsize {ViT-B} & \scriptsize VQKD-B~\cite{beitv2} & \scriptsize {IN-1K} & \scriptsize {1600} (1M-step) & \scriptsize {90 epochs, 224\suptext{2}} & \scriptsize 86.5 \ph{($*$)} \\
        \rpink
        \scriptsize \trv-B & \scriptsize \evaone-CLIP~\cite{eva} & \scriptsize {IN-21K} & \scriptsize {150} (1M-step) & \scriptsize {\xmark} & \scriptsize 87.0 \phpink{($*$)} \\
    \end{tabular}
    \vspace{-1.em}
    \caption{\textbf{MIM target representations.} When pre-trained with sufficient compute budgets and data, learning from a giant \evaone-CLIP can bring about considerable performance improvement compared with smaller CLIP teachers.}
    \vspace{-1.5em}
    \label{tab: mim target representations}
\end{table}

Although the inner-block micro architecture of plain ViT has continuously evolved since its inception in the year 2020~\cite{rpe,cait}, we notice that some significant architectural advances in language models have not yet been explored in the context of visual representation learning.
These include gated linear unit~\cite{dauphin2017language,shazeer2020glu} with sigmoid linear unit (SiLU)~\cite{hendrycks2016gaussian} / swich activation~\cite{ramachandran2017searching} (\textbf{SwiGLU}) as the feedforward network, \textbf{sub-LN}~\cite{ln,wang2022foundation} as the normalization layer, and 2D rotary position embedding (\textbf{RoPE})~\cite{su2021roformer} for positional information injection.

In~\tblref{tab: from vit to trv} we conduct a series of pilot experiments studying these architectural modifications\footnote{More technical details can be found in the Appendix. All these modifications do not bring additional parameters as well as FLOPs.}.
The pretext task is to regress the masked-out \evaone-CLIP vision features conditioned on visible image patches using IN-1K training images for 300 epochs, and the evaluation is done by fine-tuning the pre-trained base-sized models on IN-1K.
Starting with the baseline ViT configurations used in the original BEiT series pre-training~\cite{bao2021beit,beitv2,beit3} {\footnotesize \demphinline{($*$ in~\tblref{tab: from vit to trv})}}, we progressively refine the model design and make the following observations: 
(i) The performance of SwiGLU FFN is mediocre with the random weight initialization method used in BEiT, but works quite well with $\mathtt{xavier}$ $\mathtt{normal}$ weight initialization~\cite{xnorm} {\footnotesize \demphinline{(+1.1)}}.
(ii) sub-LN slightly improves the performance compared with pre-LN {\footnotesize \demphinline{(+0.2)}}. 
(iii) 2D RoPE can improve the performance {\footnotesize \demphinline{(+0.4)}}, while the standard relative position embedding~\cite{rpe,bao2021beit,beitv2} suffers from unstable pre-training with other configurations unchanged.

The final model configuration {\footnotesize \demphinline{($\dag$ in~\tblref{tab: from vit to trv})}}, called \textbf{\evared{Transform Vision}} (\trv, \figref{fig: vit and trv}b), aligns with the model architecture in current leading language models~\cite{chowdhery2022palm}, and achieves a favorable accuracy with an overall improvement of 1.6 points compared to the original configurations {\footnotesize \demphinline{(\ie, from 84.0 to 85.6)}}, with one \textit{caveat} that will be described next.

\begin{table}[!t]
\vspace{-2.5em}
    \centering
    \tablestyle{2pt}{1.2}
    \begin{tabular}{x{35}|x{55}x{45}|x{35}x{35}}
        & {\scriptsize{IN-21K}} & {\scriptsize{IN-1K}} ft & {\scriptsize{IN-1K}} ft & \demphs{{\scriptsize{IN-V2}} ft} \\ 
        method & intermed. ft? & img size & top-1 acc. & \demphs{top-1 acc.} \\
        \shline
        & \scriptsize \xmark & \scriptsize 196\suptext{2} & \scriptsize 87.0 & \scriptsize \demphs{77.6} \\
        \scriptsize \eva-B & \scriptsize \xmark & \scriptsize 448\suptext{2} & \scriptsize 88.3 & \scriptsize \demphs{79.5} \\
        & \scriptsize 40 epochs, 448\suptext{2} & \scriptsize 448\suptext{2} & \evadefault{\scriptsize \textbf{88.6}} & \evadefault{\scriptsize \demphs{\textbf{79.8}}} \\
        \hline
        & \scriptsize \xmark & \scriptsize 196\suptext{2} & \scriptsize 88.9 & \scriptsize \demphs{80.7} \\
        \scriptsize \eva-L & \scriptsize \xmark & \scriptsize 448\suptext{2} & \scriptsize 89.6 & \scriptsize \demphs{82.3} \\
        & \scriptsize 30 epochs, 448\suptext{2} & \scriptsize 448\suptext{2} & \evadefault{\scriptsize \textbf{90.0}} & \evadefault{\scriptsize \demphs{\textbf{82.4}}} \\
    \end{tabular}
    \vspace{-1.em}
    \caption{\textbf{More scaling can further boost the performance.} Pre-training and architectural configurations are detailed in~\tblref{tab: pt and arch configs}. ``IN-V2'' refers to ImageNet-V2~\cite{recht2019imagenetv2}.}
    \vspace{-1.5em}
    \label{tab: more scaling, better results}
\end{table}

\subsection{Pre-training Strategy}
\label{sec: pre-training strategy}

In the previous section, we choose to use features from a giant CLIP vision encoder with one billion parameters as the target representation for our MIM pretext task.
However, we have not yet explained the rationale behind this choice.
Although similar pre-training strategies have been well-studied in recent literature~\cite{wei2022mvp,hou2022milan,eva,dBOT,caev2} and shown to be effective, they typically use vision features from much smaller CLIP models.
Choosing the 1B-parameter \evaone-CLIP is based on our assumption that larger CLIP will provide more robust and transferable target representations for MIM, and will ultimately lead to better pre-trained models.
In~\tblref{tab: mim target representations}, we study the impact of target representations produced by different-sized CLIPs.

\paragraph{A caveat from a crash course.}
At first glance, compared with the smaller VQKD-B~\cite{beitv2} and CLIP-B~\cite{clip} as MIM teachers, the accuracy \textit{degenerate} {\footnotesize \demphinline{(\ie, from 85.0 to 84.0)}} with \evaone-CLIP target when the students are base-sized plain ViT in ~\cite{dosovitskiy2020vit,bao2021beit} with 300 epochs pre-training on IN-1K {\footnotesize \demphinline{($*$ in~\tblref{tab: from vit to trv} and~\tblref{tab: mim target representations})}}.
The architectural modifications from \trv compensate for this to some extent, resulting in a modest total improvement of 0.6-point {\footnotesize \demphinline{($\dag$ in~\tblref{tab: from vit to trv} and~\tblref{tab: mim target representations})}}.

We conjecture that as the teacher becomes stronger, it becomes harder for the students to learn robust and transferable representations in a crash course.
Consequently, more extensive pre-training is required for the students to fully master the teacher's knowledge.
As we extend the pre-training schedule to 1600 epochs {\footnotesize \demphinline{(\app 1M steps)}}, \trv with \evaone-CLIP as the MIM teacher yields a 1.3-point non-trivial improvement over BEiTv2~\cite{beitv2}.
Furthermore, with 150 epochs {\footnotesize \demphinline{(\app 1M steps)}} pure MIM pre-training on ImageNet-21K (IN-21K, 14.2M images)~\cite{deng2009imagenet}, our base-sized \trv achieves 87.0 top-1 accuracy, even outperform BEiTv2 with 1600 epochs {\footnotesize \demphinline{(\app 1M steps)}} MIM pre-training on IN-1K \textit{plus an additional 90 epochs intermediate fine-tuning on IN-21K with labels}.

\begin{table*}[!t]
\vspace{-1.em}
    \centering
    \tablestyle{4pt}{1.2}
    \begin{tabular}{l|ccc|cccccc|cc}
        & \multicolumn{3}{c|}{\textbf{MIM pre-training settings}} & \multicolumn{6}{c|}{\textbf{macro arch configs} (refer to \tblref{tab: from vit to trv} \& \figref{fig: vit and trv}b for micro arch)} & enc. & FLOPs \\
        model & teacher & pt data & pt epochs & patch size & depth & width & attn heads & FFN type & FFN hidden dim & \#params & (\#tokens = 196) \\
        \shline
        \scriptsize \eva-Ti & \scriptsize \evaone-CLIP & \scriptsize IN-21K (14M) & 240 & 14$\times$14 & 12 & 192 & 3 & SwiGLU & 512 & 6M & 1.3G \\
        \scriptsize \eva-S & \scriptsize \evaone-CLIP & \scriptsize IN-21K (14M) & 240 & 14$\times$14 & 12 & 384 & 6 & SwiGLU & 1024 & 22M & 4.6G \\
        \scriptsize \eva-B & \scriptsize \evaone-CLIP & \scriptsize IN-21K (14M) & 150 & 14$\times$14 & 12 & 768 & 12 & SwiGLU & 2048 & 86M & 18G \\
        \scriptsize \eva-L & \scriptsize \evaone-CLIP & \scriptsize Merged-38M & 56 & 14$\times$14 & 24 & 1024 & 16 & SwiGLU & 2730 & 304M & 62G \\
    \end{tabular}
    \vspace{-.5em}
    \caption{\textbf{Summary of MIM pre-training settings and architecture configurations.}}
    \vspace{-0.75em}
    \label{tab: pt and arch configs}
\end{table*}

Ulteriorly, in~\tblref{tab: more scaling, better results} we show that scaling model size, resolution as well as injecting labels via intermediate fine-tuning can further boost the performance, reaching up to 90.0 top-1 accuracy on IN-1K with only a 304M-parameter \eva.
Notably, our \textit{pure} MIM pre-trained representations can achieve very competitive performance \textit{without} additional intermediate fine-tuning.

From now on, we denote \trv with sufficient MIM pre-training from \evaone-CLIP as \eva.
In the rest of this section, we present some technical details of MIM pre-training before go into the performance evaluation in~\sref{sec: experiments}.

\paragraph{Model variants and architectures.}
We provide four variants, \ie, \eva-Ti {\footnotesize \demphinline{(6M)}}, -S {\footnotesize \demphinline{(22M)}}, -B {\footnotesize \demphinline{(86M)}} and -L {\footnotesize \demphinline{(304M)}}, as detailed in~\tblref{tab: pt and arch configs}. 
The marco architecture {\footnotesize \demphinline{(\eg, model depth, width, \#head)}} of \eva variants follows the canonical plain ViT configurations in~\cite{touvron2021training,dosovitskiy2020vit}. 
The inner-block modifications are detailed in~\sref{sec: architecture}.

\paragraph{Pre-training objective} is similar to \evaone~\cite{eva}, which is to regress the masked-out image-text aligned vision features conditioned on visible image patches only.
We corrupt the input patches with $\mathtt{[MASK]}$ tokens, and we use block-wise masking with a masking ratio of 40\% following~\cite{bao2021beit,eva}.
The target representation for MIM pre-training is from the publicly accessible \evaone-CLIP~\cite{eva} vision tower with one billion parameters.
The output feature of \eva is first normalized~\cite{ln} and then projected to the same dimension as the \evaone-CLIP's vision feature via a linear layer.
We use negative cosine similarity as the loss function.

\paragraph{Pre-training data.}
For \eva-Ti, -S and -B, we use images from IN-21K~\cite{deng2009imagenet} for pre-training.
For \eva-L, we use a merged dataset consisting of IN-21K, CC12M~\cite{CC12M}, CC3M~\cite{CC3M}, COCO~\cite{lin2014coco}, ADE20K~\cite{zhou2018ade}, Object365~\cite{o365} and OpenImages~\cite{openimages}.
For CC12M and CC3M, we only use the image data without captions.
For COCO and ADE20K, we only use the training set images.
The merged dataset for pre-training \eva-L has 38 million images in total {\footnotesize \demphinline{(denoted as Merged-38M)}}.
All these datasets are publicly accessible.

\paragraph{Hyper-parameters} generally follow the BEiT series~\cite{bao2021beit,beitv2,beit3}. 
The optimizer is Adam~\cite{adam} with decoupled weight decay~\cite{Loshchilov2019adamw} / $\beta_2$ of 0.05 / 0.98~\cite{liu2019roberta}.
The peak learning rate / batch size is 3e-3 / 4k for tiny- and small-sized models, and 1.5e-3 / 2k for base- and large-sized models.
We train tiny- and small-sized models for \app0.8M steps, and base- and large-sized models for \app1M steps.

\paragraph{Implementation.}
The pre-training code is based on the open-sourced \evaone implementation~\cite{pytorch,eva,EVA_code_models}.
We adopt $\mathtt{DeepSpeed}$~\cite{rasley2020deepspeed} with ZeRO stage-0 / -1 optimizer and $\mathtt{fp16}$ precision with dynamic loss scaling~\cite{rajbhandari2020zero}.
All MHSA operations are accelerated by $\mathtt{xFormers}$~\cite{xFormers2022}.
Although our MIM teacher comes with one billion parameters, the wall-clock pre-training time is \app10\% shorter than the official BEiT series implementations~\cite{bao2021beit,beitv2}.

\section{Experiments and Evaluation}
\label{sec: experiments}

In this section, we present a comprehensive evaluation of our approach on representative vision tasks and benchmarks, including image classification in~\sref{sec: image classification}, contrastive image-text pre-training (CLIP) with zero-shot evaluation in~\sref{sec: clip}, object detection \& instance segmentation in~\sref{sec: od}, and semantic segmentation in~\sref{sec: seg}. 
We conduct experiments mainly using base-sized {\footnotesize \demphinline{(86M)}} and large-sized {\footnotesize \demphinline{(304M)}} pre-trained representations. 
Our results demonstrate that \eva is capable of outperforming larger counterparts and achieving state-of-the-art performance without or with only minimal additional intermediate fine-tuning. 
Additional details and results can be found in the Appendix.

\subsection{Image Classification}
\label{sec: image classification}

\paragraph{Datasets.}
For image classification, we mainly evaluate \eva on IN-1K~\cite{russakovsky2015imagenet}.
We also evaluate the robustness \& generalization capability of \eva along with our training settings \& hyper-parameters using some IN-1K validation set variants, including ImageNet-V2 matched frequency {\footnotesize{\demphinline{(IN-V2)}}}~\cite{inv2}, ImageNet-ReaL {\footnotesize{\demphinline{(IN-ReaL)}}}~\cite{inreal}, ImageNet-Adversarial {\footnotesize{\demphinline{(IN-Adv.)}}}~\cite{inadv}, ImageNet-Rendition {\footnotesize{\demphinline{(IN-Ren.)}}}~\cite{inren}, ImageNet-Sketch {\footnotesize{\demphinline{(IN-Ske.)}}}~\cite{inske}, as well as ObjectNet {\footnotesize{\demphinline{(ObjNet)}}}~\cite{objectnet}, following the settings in~\cite{he2021masked,eva}.

\begin{table*}[!b] 
    \centering
    \tablestyle{2.5pt}{1.2}
    \begin{tabular}{l|rc|ccccccc|c|r}
        method & \#params & data & \scriptsize IN-1K~\cite{russakovsky2015imagenet} & \scriptsize IN-V2~\cite{inv2} & \scriptsize IN-ReaL~\cite{inreal} & \scriptsize IN-Adv.~\cite{inadv} & \scriptsize IN-Ren.~\cite{inren} & \scriptsize IN-Ske.~\cite{inske} & \scriptsize \demphs{ObjNet}~\cite{objectnet} & avg. & $\Delta$\scriptsize{$\downarrow$} \\
        \shline
        \multicolumn{12}{c}{\scriptsize{(a) comparisons with SOTA \textbf{base}-sized models (86M)}} \\
        \hline
        \scriptsize LAION-ViT-CLIP-B$\dag$~\cite{laionftvitb} & 86M\ph{-} & \scriptsize LAION-2B \& IN-21K & 87.2 & 77.8 & 90.2 & 59.2 & 66.2 & 53.5 & \demphs{-} & 72.4 & 14.8 \\
        \scriptsize DeiT-III-H~\cite{deit3} & 632M\ph{-} & \scriptsize IN-21K & 87.2 & 79.2 & 90.2 & 70.2 & 70.8 & 55.8 & \demphs{-} & 75.6 & 11.6 \\
        \rpink
        \scriptsize \eva-B & \textbf{86M}\phpink{-} & \scriptsize IN-21K & \textbf{88.6} & \textbf{79.8} & \textbf{90.8} & \textbf{78.1} & \textbf{76.8} & \textbf{57.7} & \demphs{55.3} & \textbf{78.6} & \textbf{10.0} \\
        \shline
        \multicolumn{12}{c}{\scriptsize{(b) comparisons with \textbf{larger} SOTA models}} \\
        \hline
        \scriptsize LAION-ViT-CLIP-H$\dag$~\cite{laionftvith} & 632M\ph{-} & \scriptsize LAION-2B \& IN-21K & 88.6 & 79.5 & 90.5 & 74.2 & 83.1 & 65.3 & \demphs{-} & 80.2 & 8.4 \\
        \scriptsize \evaone~\cite{eva} (prev. best) & 1011M\ph{-} & \scriptsize Merged-30M & 89.6 & 81.6 & 90.8 & 86.2 & 88.3 & 67.7 & \demphs{60.9} & 84.0 & 5.6 \\
        \rpink
        \scriptsize \eva-L & \textbf{304M}\phpink{-} & \scriptsize Merged-38M & \textbf{90.0} & \textbf{82.4} & \textbf{91.1} & \textbf{87.7} & \textbf{89.9} & \textbf{70.1} & \demphs{62.8} & \textbf{85.2} & \textbf{4.8} \\
        \end{tabular}
        \vspace{-1.em}
        \caption{\textbf{Robustness \& generalization capability evaluation on IN-1K variants.} All these models are first fine-tuned on the original IN-1K training set and then evaluated on different val sets using the \textit{same} fine-tuned model \textit{without any specialized fine-tuning}. ``avg.'': the averaged top-1 accuracy on different IN-1K val set variants ({\footnotesize{\demphinline{\ie, IN-\{1K, V2, ReaL, Adv., Ren., Ske.\}, excluding ObjNet}}}). ``{$\Delta$\scriptsize{$\downarrow$}}'': The gap between the averaged top-1 accuracy of val set variants and the original IN-1K validation set top-1 accuracy (the lower the better). \\ {\footnotesize \demphinline{``$\dag$'': fine-tuned CLIP vision encoder}}}
        \label{tab: cls rob and gen}
\end{table*}

\paragraph{Training settings.}
To fully unleash the potential of \eva, we optionally perform intermediate fine-tuning following~\cite{bao2021beit,beitv2} for base- / large-sized model on IN-21K~\cite{deng2009imagenet} for 40 / 30 epochs in~\tblref{tab: 1k cls}.
The final IN-1K fine-tuning for all-sized models {\footnotesize{\demphinline{(including \eva-Ti and -S)}}} can be done without using strong regularization such as cutmix~\cite{yun2019cutmix}, mixup~\cite{zhang2017mixup} and random erasing~\cite{zhong2020random}.
In the Appendix, we show that our pre-trained representations are robust enough that can be fine-tuned using various numerical precisions {\footnotesize \demphinline{(\eg, $\mathtt{fp16}$ and $\mathtt{bf16}$)}} and optimizers {\footnotesize \demphinline{(\eg, Lion~\cite{lion}, AdamW~\cite{adam,Loshchilov2019adamw}, and SGD~\cite{sgd})}}. Remarkably, the fine-tuning can be done even using the SGD optimizer with only 0.1-point performance drop.

\paragraph{IN-1K results (\eva-B \& -L).}
\tblref{tab: 1k cls} compares \eva with some state-of-the-art models on IN-1K val set.
Our base-sized model, trained with ImageNet data only, outperforms several strong competitors and achieves the same performance with a ViT-B distilled from a 4B-parameter teacher using large-scale in-house training data~\cite{vit22b}.
Furthermore, \eva-L with only 304M-parameter can achieve a phenomenal 90.0 fine-tuning top-1 accuracy, outperforms several state-of-the-art larger models trained with more (often publicly inaccessible) data, \textit{including its fine-tuned \evaone-CLIP MIM teacher}, which distinguishes MIM from knowledge distillation~\cite{hinton2015distilling}.

\begin{table}[!t]
    \centering
    \tablestyle{2.0pt}{1.2}
    \begin{tabular}{l|rcc|c}
            & & extra & crop & \scriptsize IN-1K \\
            method & \#params & labeled data & size & top-1 \\
            \shline
            \multicolumn{5}{c}{\scriptsize{(a) comparisons with SOTA \textbf{base}-sized models (86M)}} \\
            \hline
            \scriptsize LAION-ViT-CLIP-B$\dag$~\cite{laionftvitb} & \scriptsize 86M & \scriptsize{LAION-2B \& IN-21K} & \scriptsize 384\suptext{2} & \scriptsize 87.2 \\
            \scriptsize BEiTv2-B~\cite{beitv2} & \scriptsize 86M & \scriptsize{IN-21K (14M)} & \scriptsize 384\suptext{2} & \scriptsize 87.5 \\
            \scriptsize ViT-B \alambic ViT-22B-JFT-4B~\cite{vit22b} & \scriptsize 86M & \scriptsize{JFT-4B} & \scriptsize 384\suptext{2} & \scriptsize 88.6 \\
            \rpink
            \scriptsize \eva-B & \scriptsize 86M & \scriptsize{IN-21K (14M)} & \scriptsize 448\suptext{2} & \scriptsize \textbf{88.6} \\
            \shline
            \multicolumn{5}{c}{\scriptsize{(b) comparisons with \textbf{larger} SOTA models}} \\
            \hline
            \scriptsize LAION-ViT-CLIP-L$\dag$~\cite{laionftvitl} & \scriptsize 304M & \scriptsize{LAION-2B \& IN-21K} & \scriptsize 336\suptext{2} & \scriptsize 88.2 \\
            \scriptsize FD-CLIP-L~\cite{wei2022featdistill} & \scriptsize 304M & \scriptsize{IN-21K (14M)} & \scriptsize 336\suptext{2} & \scriptsize 89.0 \\
            \scriptsize BEiTv2-L~\cite{beitv2} & \scriptsize 304M & \scriptsize{IN-21K (14M)} & \scriptsize 384\suptext{2} & \scriptsize 89.2 \\
            \scriptsize ViT-L \alambic ViT-22B-JFT-4B~\cite{vit22b} & \scriptsize 304M & \scriptsize{JFT-4B} & \scriptsize 384\suptext{2} & \scriptsize 89.6 \\
            \rpink
            \scriptsize \eva-L & \scriptsize 304M & \scriptsize{IN-21K (14M)} & \scriptsize 448\suptext{2} & \scriptsize \textbf{90.0} \\
            \scriptsize \demphs{InternImage-H~\cite{internimage}} & \scriptsize \demphs{$\appp$1080M} & \scriptsize{\demphs{427M img-txt \& IN-21K}} & \scriptsize \demphs{640\suptext{2}} & \scriptsize \demphs{89.2} \\
            \scriptsize \demphs{\evaone-CLIP$\dag$~\cite{eva}} & \scriptsize \demphs{1011M} & \scriptsize{\demphs{IN-21K (14M)}} & \scriptsize \demphs{336\suptext{2}} & \scriptsize \demphs{89.5} \\
            \scriptsize \demphs{BEiT-3~\cite{beit3}} & \scriptsize \demphs{$\appp$1900M} & \scriptsize{\demphs{400M img-txt \& IN-21K}} & \scriptsize \demphs{336\suptext{2}} & \scriptsize \demphs{89.6} \\
            \scriptsize \demphs{\evaone~\cite{eva}} & \scriptsize \demphs{1011M} & \scriptsize{\demphs{IN-21K (14M)}} & \scriptsize \demphs{560\suptext{2}} & \scriptsize \demphs{89.7} \\
            \scriptsize \demphs{RevCol-H~\cite{revcol}} & \scriptsize \demphs{2158M} & \scriptsize{\demphs{168M (semi sup.)}} & \scriptsize \demphs{640\suptext{2}} & \scriptsize \demphs{90.0} \\
        \end{tabular}
        \vspace{-1.em}
        \caption{\textbf{\eva-B and \eva-L image classification performance on IN-1K val set.} Using only publicly accessible data, \eva creates phenomenal results with affordable model size. \\ {\footnotesize \demphinline{``$\dag$'': fine-tuned CLIP vision encoder. \quad ``\alambic'': model distillation~\cite{hinton2015distilling,beyer2022knowledge}.}}}
        \vspace{-1.5em}
        \label{tab: 1k cls}
\end{table}

\paragraph{IN-1K results (\eva-Ti \& -S).}
It is commonly believed that plain ViTs perform mediocrely due to the lack of inductive biases in light-weight settings.
However, compared with specialized light-weight networks with strong visual structure prior in~\tblref{tab: ti and s 1k cls}, \eva as a plain ViT variant equipped with extensive MIM pre-training can trump inductive biases, and achieve favorable performance with tiny and small models.

\begin{table}[!t]
    \centering
    \tablestyle{4.8pt}{1.2}
    \begin{tabular}{l|rccc|c}
        & & {\scriptsize{IN-1K}} ft & & {\scriptsize{IN-21K}} & {\scriptsize{IN-1K}} \\
        method & \#params & img size & \scriptsize FLOPs & label? & top-1 \\
        \shline
        \multicolumn{6}{c}{\scriptsize{(a) model size: 5M$\appp$10M}} \\
        \hline
        \scriptsize MobileViTv3-1.0~\cite{wadekar2022mobilevitv3} & 5.1M & 384\suptext{2} & 4.2G & \xmark & 79.7 \\ 
        \scriptsize MobileViTv2-1.5~\cite{mehta2022separable} & 10.6M & 256\suptext{2} & 4.0G & \xmark & 80.4 \\ 
        \rpink
        \scriptsize \eva-Ti & 5.7M & 336\suptext{2} & 4.8G & \xmark & \textbf{80.7} \\
        \shline
        \multicolumn{6}{c}{\scriptsize{(b) model size: 20M$\appp$30M}} \\
        \hline
        \scriptsize DeiT-III-S~\cite{deit3} & 22M & 384\suptext{2} & 16G & \cmark & 84.8 \\
        \scriptsize ConvNeXt V2-T~\cite{woo2023convnext} & 29M & 384\suptext{2} & 13G & \cmark & 85.1 \\
        \scriptsize MOAT-0~\cite{yang2022moat} & 28M & 384\suptext{2} & 18G & \cmark & 85.7 \\
        \rpink
        \scriptsize \eva-S & 22M & 336\suptext{2} & 16G & \xmark & \textbf{85.8} \\
        \scriptsize \demphs{BEiTv2-B}~\cite{beitv2} & \demphs{86M} & \demphs{224\suptext{2}} & \demphs{18G} & \demphs{\xmark} & \demphs{85.5} \\
    \end{tabular}
    \vspace{-1.em}
    \caption{\textbf{\eva-Ti and \eva-S image classification performance on IN-1K val set.} \eva with fewer inductive biases but sufficient MIM pre-training is performant in light-weight settings.}
    \vspace{-1.5em}
    \label{tab: ti and s 1k cls}
\end{table}

\paragraph{Robustness evaluation.}
We evaluate the robustness and generalization capability of \eva on several IN-1K val set variants.
Following the evaluation procedure in~\cite{he2021masked,eva}, all these models are first fine-tuned on the original IN-1K training set, and then directly evaluated on different val sets using the \textit{same} fine-tuned model \textit{without further hyper-parameter selection and specialized fine-tuning}.

In \tblref{tab: cls rob and gen}, we compare \eva with some top open-sourced models.
\eva is the most competitive one in terms of top-1 accuracies.
Besides the absolute performance, we also care about whether a model along with its training settings biases towards the original validation set and generalizes well on others.
From this perspective, \eva not only achieves the highest averaged accuracy, but also has the smallest performance gap {\footnotesize \demphinline{(as measured by the difference between the averaged accuracy of val set variants and the original IN-1K val set accuracy)}}, which reflects the excellent robustness and generalization ability of \eva.

\begin{table*}[!b]
\centering
\tablestyle{1.4pt}{1.2}
    \begin{tabular}{r|cccccc|ccccccccccccccccccccc|c}
        &
        \rotatebox[origin=l]{90}{\scriptsize{ImageNet-1K~\cite{russakovsky2015imagenet}}} &
        \rotatebox[origin=l]{90}{\scriptsize{ImageNet-V2~\cite{inv2}}} &
        \rotatebox[origin=l]{90}{\scriptsize{ImageNet-Adv.~\cite{inadv}}} &
        \rotatebox[origin=l]{90}{\scriptsize{ImageNet-Ren.~\cite{inren}}} &
        \rotatebox[origin=l]{90}{\scriptsize{ImageNet-Ske.~\cite{inske}}} &
        \rotatebox[origin=l]{90}{\scriptsize{ObjectNet~\cite{objectnet}}} &
        \rotatebox[origin=l]{90}{\scriptsize{CIFAR-10~\cite{cifar}}} &
        \rotatebox[origin=l]{90}{\scriptsize{CIFAR-100~\cite{cifar}}} & 
        \rotatebox[origin=l]{90}{\scriptsize{MNIST~\cite{lecun1998gradient}}} & 
        \rotatebox[origin=l]{90}{\scriptsize{Caltech-101~\cite{fei2004learning}}} & 
        \rotatebox[origin=l]{90}{\scriptsize{SUN397~\cite{xiao2010sun}}} & 
        \rotatebox[origin=l]{90}{\scriptsize{FGVC Aircraft~\cite{maji2013fine}}} & 
        \rotatebox[origin=l]{90}{\scriptsize{Country-211~\cite{clip}}} & 
        \rotatebox[origin=l]{90}{\scriptsize{Stanford Cars~\cite{krause20133d}}} &
        \rotatebox[origin=l]{90}{\scriptsize{Birdsnap~\cite{berg2014birdsnap}}} & 
        \rotatebox[origin=l]{90}{\scriptsize{DTD~\cite{cimpoi14describing}}} & 
        \rotatebox[origin=l]{90}{\scriptsize{Eurosat~\cite{helber2019eurosat}}} & 
        \rotatebox[origin=l]{90}{\scriptsize{FER2013~\cite{goodfellow2013challenges}}} & 
        \rotatebox[origin=l]{90}{\scriptsize{Flowers-102~\cite{nilsback2008automated}}} & 
        \rotatebox[origin=l]{90}{\scriptsize{Food-101~\cite{bossard2014food}}} & 
        \rotatebox[origin=l]{90}{\scriptsize{GTSRB~\cite{stallkamp2012man}}} & 
        \rotatebox[origin=l]{90}{\scriptsize{PCam~\cite{veeling2018rotation}}} & 
        \rotatebox[origin=l]{90}{\scriptsize{Pets~\cite{parkhi12a}}} & 
        \rotatebox[origin=l]{90}{\scriptsize{Rendered SST2~\cite{clip}}} & 
        \rotatebox[origin=l]{90}{\scriptsize{Resisc45~\cite{cheng2017remote}}} & 
        \rotatebox[origin=l]{90}{\scriptsize{STL10~\cite{coates2011analysis}}} & 
        \rotatebox[origin=l]{90}{\scriptsize{VOC2017~\cite{everingham2015pascal}}} &
        \rotatebox[origin=l]{90}{\ph{.}\textbf{avg. top-1 acc.}}
        \\
        \shline
        \multicolumn{29}{c}{\scriptsize (a) comparisons with CLIP-\textbf{Base} baselines} \\
        \hline
        \scriptsize OpenAI CLIP-B/16\ph{+} & \scriptsize 68.3 & \scriptsize 61.9 & \scriptsize 50.0 & \scriptsize 77.7 & \scriptsize 48.2 & \scriptsize 55.3 & \scriptsize 90.8 & \scriptsize 67.0 & \scriptsize \textbf{51.6} & \scriptsize 84.7 & \scriptsize 64.4 & \scriptsize 24.4 & \scriptsize \textbf{22.8} & \scriptsize 64.8 & \scriptsize 34.5 & \scriptsize 44.7 & \scriptsize 55.0 & \scriptsize 46.2 & \scriptsize 71.3 & \scriptsize 88.8 & \scriptsize 43.5 & \scriptsize 50.7 & \scriptsize 89.1 & \scriptsize \textbf{60.8} & \scriptsize 59.1 & \scriptsize 98.3 & \scriptsize 78.3 & \scriptsize 61.2 \\
        \rpink
        \scriptsize \eva-CLIP-B/16\phpink{+} & \scriptsize \textbf{74.7} & \scriptsize \textbf{67.0} & \scriptsize \textbf{54.1} & \scriptsize \textbf{82.5} & \scriptsize \textbf{57.7} & \scriptsize \textbf{62.3} & \scriptsize \textbf{98.4} & \scriptsize \textbf{87.7} & \scriptsize 47.9 & \scriptsize \textbf{86.3} & \scriptsize \textbf{70.7} & \scriptsize \textbf{24.8} & \scriptsize 21.4 & \scriptsize \textbf{78.6} & \scriptsize \textbf{37.7} & \scriptsize \textbf{53.1} & \scriptsize \textbf{67.0} & \scriptsize \textbf{51.2} & \scriptsize \textbf{75.9} & \scriptsize \textbf{89.4} & \scriptsize \textbf{46.3} & \scriptsize \textbf{50.9} & \scriptsize \textbf{92.2} & \scriptsize 54.1 & \scriptsize \textbf{60.7} & \scriptsize \textbf{99.5} & \scriptsize \textbf{80.2} & \scriptsize \textbf{65.6} \\
        \shline
        \multicolumn{29}{c}{\scriptsize (b) comparisons with \textbf{larger} CLIP models} \\
        \hline
        \scriptsize OpenAI CLIP-L/14+ & \scriptsize 76.6 & \scriptsize 70.9 & \scriptsize 77.5 & \scriptsize 89.0 & \scriptsize 61.0 & \scriptsize 72.0 & \scriptsize 94.9 & \scriptsize 74.4 & \scriptsize \textbf{79.0} & \scriptsize 87.2 & \scriptsize 68.7 & \scriptsize 33.4 & \scriptsize \textbf{34.5} & \scriptsize 79.3 & \scriptsize 41.0 & \scriptsize 56.0 & \scriptsize 61.5 & \scriptsize 49.1 & \scriptsize 78.6 & \scriptsize 93.9 & \scriptsize 52.4 & \scriptsize \textbf{60.8} & \scriptsize 93.8 & \scriptsize \textbf{70.7} & \scriptsize 65.4 & \scriptsize 99.4 & \scriptsize 78.1 & \scriptsize 70.3 \\
        \scriptsize OpenCLIP-H/14\ph{+} & \scriptsize 78.0 & \scriptsize 70.8 & \scriptsize 59.2 & \scriptsize 89.3 & \scriptsize 66.6 & \scriptsize 69.7 & \scriptsize 97.4 & \scriptsize 84.7 & \scriptsize 72.9 & \scriptsize 85.0 & \scriptsize \textbf{75.2} & \scriptsize \textbf{42.8} & \scriptsize 30.0 & \scriptsize \textbf{93.5} & \scriptsize \textbf{52.9} & \scriptsize \textbf{67.8} & \scriptsize \textbf{72.7} & \scriptsize \textbf{52.0} & \scriptsize \textbf{80.1} & \scriptsize 92.7 & \scriptsize \textbf{58.4} & \scriptsize 54.2 & \scriptsize \textbf{94.5} & \scriptsize 64.3 & \scriptsize \textbf{70.5} & \scriptsize 98.5 & \scriptsize 77.7 & \scriptsize 72.3 \\
        \rpink
        \scriptsize \eva-CLIP-L/14+ & \scriptsize{\textbf{80.4}} & \scriptsize{\textbf{73.8}} & \scriptsize{\textbf{82.9}} & \scriptsize{\textbf{93.2}} & \scriptsize{\textbf{68.9}} & \scriptsize{\textbf{78.4}} & \scriptsize{\textbf{98.9}} & \scriptsize{\textbf{89.8}} & \scriptsize{64.3} & \scriptsize{\textbf{89.5}} & \scriptsize{74.8} & \scriptsize{37.5} & \scriptsize{33.6} & \scriptsize{91.6} & \scriptsize 45.8 & \scriptsize 64.5 & \scriptsize 71.4 & \scriptsize 51.0 & \scriptsize 77.2 & \scriptsize \textbf{94.2} & \scriptsize 57.6 & \scriptsize 54.9 & \scriptsize 94.2 & \scriptsize 64.6 & \scriptsize 69.8 & \scriptsize \textbf{99.7} & \scriptsize \textbf{82.7} & \scriptsize \textbf{73.5} \\
        \end{tabular}
\vspace{-.5em}
\caption{\textbf{Summary of \eva-CLIP zero-shot image classification performance on 27 datasets.}}
\label{tab: clip zs img cls 27}
\end{table*}

\subsection{Contrastive Language-Image Pre-training \\ and Zero-shot Evaluation}
\label{sec: clip}
Contrastive Language-Image Pre-trained (CLIP) model is a kind of foundation model that aligns vision and natural language through contrastive image-text pre-training~\cite{clip}.
Its impact on the field of representation learning has been significant, making it a powerful engine for both recognition and generation tasks, as well as uni-modal and multi-modal applications~\cite{dalle2,eva,blip2,laion5b}.

In this section, we thoroughly demonstrate the efficacy of initializing \eva as the CLIP vision encoder following the settings in~\cite{eva}.
The resulting model, referred to as \eva-CLIP, significantly improves zero-shot performance, sample efficiency, and training speed.

\begin{table}[!t]
    \centering
    \tablestyle{1.7pt}{1.2}
    \begin{tabular}{r|rc|r|c|c}
        & \#params\ph{+} & & & \scriptsize img & \scriptsize IN-1K \\
        method{\scriptsize{\ph{+}}} & \scriptsize (img+text)\ph{.} & \scriptsize precision & \scriptsize \ph{.}dataset \& samples\ph{.} & \scriptsize size & \scriptsize zs top-1 \\
        \shline
        \multicolumn{6}{c}{\scriptsize (a) comparisons with CLIP-\textbf{Base} baselines} \\
        \hline
        \scriptsize OpenAI CLIP-B/16\ph{+} & \scriptsize 86M+63M & \scriptsize $\mathtt{fp16}$ & \scriptsize WIT-400M \& 13B & \scriptsize 224\suptext{2} & \scriptsize 68.3 \\
        \scriptsize OpenCLIP-B/16\ph{+} & \scriptsize 86M+63M & \scriptsize $\mathtt{bf16}$ & \scriptsize LAION-2B \& 34B & \scriptsize 224\suptext{2} & \scriptsize 70.2 \\
        \rpink
        \scriptsize \eva-CLIP-B/16\phpink{+} & \scriptsize 86M+63M & \scriptsize $\mathtt{fp16}$ & \scriptsize Merged-2B \& \phpink{0}\textbf{8B} & \scriptsize 224\suptext{2} & \scriptsize \textbf{74.7} \\
        \shline
        \multicolumn{6}{c}{\scriptsize (b) comparisons with CLIP-\textbf{Large} baselines} \\
        \hline
        \scriptsize OpenAI CLIP-L/14\ph{+} & \scriptsize 0.3B+124M & \scriptsize $\mathtt{fp16}$ & \scriptsize WIT-400M \& 13B & \scriptsize 224\suptext{2} & \scriptsize 75.5 \\
        \scriptsize OpenCLIP-L/14\ph{+} & \scriptsize 0.3B+124M & \scriptsize $\mathtt{bf16}$ & \scriptsize LAION-2B \& 32B & \scriptsize 224\suptext{2} & \scriptsize 75.3 \\
        \rpink
        \scriptsize \eva-CLIP-L/14\phpink{+} & \scriptsize 0.3B+124M & \scriptsize $\mathtt{fp16}$ & \scriptsize Merged-2B \& \phpink{0}\textbf{4B} & \scriptsize 224\suptext{2} & \scriptsize \textbf{79.8} \\
        \shline
        \multicolumn{6}{c}{\scriptsize (c) comparisons with \textbf{larger} CLIPs trained with \textbf{more samples}} \\
        \hline
        \scriptsize OpenAI CLIP-L/14+ & \scriptsize 0.3B+124M & \scriptsize $\mathtt{fp16}$ & \scriptsize WIT-400M \& 13B & \scriptsize 336\suptext{2} & \scriptsize 76.6 \\
        \scriptsize OpenCLIP-H/14\ph{+} & \scriptsize 0.6B+354M & \scriptsize $\mathtt{bf16}$ & \scriptsize LAION-2B \& 32B & \scriptsize 224\suptext{2} & \scriptsize 78.0 \\
        \scriptsize FLIP-H/14\ph{+} & \scriptsize 0.6B+354M & \scriptsize $\mathtt{fp32}$ & \scriptsize LAION-2B \& 26B & \scriptsize 224\suptext{2} & \scriptsize 78.1 \\
        \scriptsize \evaone-CLIP-g/14\ph{+} & \scriptsize 1.0B+124M & \scriptsize $\mathtt{fp16}$ & \scriptsize LAION-0.4B \& 11B & \scriptsize 224\suptext{2} & \scriptsize 78.5 \\
        \scriptsize \alambic OpenCLIP-G/14\ph{+} & \scriptsize 1.8B+695M & \scriptsize $\mathtt{bf16}$ & \scriptsize LAION-2B \& 39B & \scriptsize 224\suptext{2} & \scriptsize 80.1 \\
        \rpink
        \scriptsize \eva-CLIP-L/14+ & \scriptsize \textbf{0.3B}+\textbf{124M} & \scriptsize $\mathtt{fp16}$ & \scriptsize Merged-2B \& \phpink{0}\textbf{6B} & \scriptsize 336\suptext{2} & \scriptsize \textbf{80.4} \\
    \end{tabular}
    \vspace{-1.em}
    \caption{\textbf{CLIP configurations \& IN-1K zero-shot performance.} \eva-CLIP achieves better performance with affordable size and fewer image-text samples. \\ {\footnotesize \demphinline{``+'': initialized from CLIP checkpoint trained with 224\suptext{2} following~\cite{clip} \\ ``\alambic'': model soups~\cite{modelsoups}}}}
    \vspace{-1.5em}
    \label{tab: clip configs and 1k zs}
\end{table}

\paragraph{CLIP configurations \& zero-shot classification.}
We present CLIP model configurations and IN-1K zero-shot accuracies in~\tblref{tab: clip configs and 1k zs}. 
To train \eva-CLIP, we merge the data from the publicly accessible LAION-2B~\cite{laion5b} and COYO-700M~\cite{kakaobrain2022coyo-700m}, which results in a dataset of 2 billion image-text pairs {\footnotesize \demphinline{(we only have \app1.6B / \app400M valid samples from LAION-2B / COYO-700M datasets)}}. 
Leveraging MIM pre-trained \eva representations, our CLIP model significantly outperforms previous approaches in IN-1K zero-shot classification, achieving an outstanding 74.7 / 80.4 top-1 accuracy with base- / large-sized models. 

In~\tblref{tab: clip zs img cls 27}, we further demonstrate the efficacy and robustness of our approach on 26 additional zero-shot classification benchmarks. 
Notably, our \eva-CLIP-L model, which only has \app1/2 of the model size and \app1/5 image-text pairs, achieves a 1.2-point non-trival averaged improvement over OpenCLIP-H.

Finally, in~\tblref{tab: clip zs video cls} we show that \eva-CLIP is also quite effective in zero-shot video recognition benchmarks.

\begin{table}[!t]
    \centering
    \tablestyle{1.5pt}{1.2}
    \begin{tabular}{r|c|x{27}x{20}x{20}x{20}|c}
        & \#params & & & & & \\
        method\ph{+} & \scriptsize (img+text) & \scriptsize UCF-101 & \scriptsize K-400 & \scriptsize K-600 & \scriptsize K-700 & \textbf{avg. acc.} \\
        \shline
        \multicolumn{7}{c}{\scriptsize{(a) comparisons with CLIP-\textbf{Base} baselines}} \\
        \hline
        \scriptsize OpenAI CLIP-B/16\ph{+} & \scriptsize 86M+63M & \scriptsize 67.1 & \scriptsize \textbf{57.6} & \scriptsize 56.5 & \scriptsize 49.3 & \scriptsize 57.6 \\
        \rpink
        \scriptsize \eva-CLIP-B/16\phpink{+} & \scriptsize 86M+63M & \scriptsize \textbf{68.6} & \scriptsize 57.4 & \scriptsize \textbf{57.0} & \scriptsize \textbf{50.0} & \scriptsize \textbf{58.3} \\
        \shline
        \multicolumn{7}{c}{\scriptsize{(b) comparisons with \textbf{larger}-sized CLIP models}} \\
        \hline
        \scriptsize OpenAI CLIP-L/14+ & \scriptsize 0.3B+124M & \scriptsize 78.1 & \scriptsize 64.9 & \scriptsize 65.0 & \scriptsize 58.5 & \scriptsize 66.6 \\
        \scriptsize OpenCLIP-H/14\ph{+} & \scriptsize 0.6B+354M & \scriptsize 78.2 & \scriptsize 63.1 & \scriptsize 63.6 & \scriptsize 56.1 & \scriptsize 65.3 \\
        \rpink
        \scriptsize \eva-CLIP-L/14+ & \scriptsize 0.3B+124M & \scriptsize \textbf{78.6} & \scriptsize \textbf{65.9} & \scriptsize \textbf{66.1} & \scriptsize \textbf{60.2} & \scriptsize \textbf{67.7} \\
    \end{tabular}
    \vspace{-1.em}
    \caption{\textbf{Zero-shot video classification performance.} Following~\cite{clip}, we report top-1 accuracy for UCF-101~\cite{ucf101}, and the mean of top-1 and top-5 accuracies for K-400~\cite{carreira2017quo}, K-600~\cite{k600} and K-700~\cite{k700} datasets.}
    \vspace{-1.5em}
    \label{tab: clip zs video cls}
\end{table}

\begin{table*}[t!]
\vspace{-.6em}
\centering
    \tablestyle{4.2pt}{1.2}
    \begin{tabular}{r|ccc|ccc|ccc|ccc|ccc}
        & & & & \multicolumn{6}{c|}{zero-shot \textbf{text} retrieval} & \multicolumn{6}{c}{zero-shot \textbf{image} retrieval} \\
        & \#params\ph{+} & & img-text & \multicolumn{3}{c|}{\scriptsize Flickr30K} & \multicolumn{3}{c|}{\scriptsize COCO} & \multicolumn{3}{c|}{\scriptsize Flickr30K} & \multicolumn{3}{c}{\scriptsize COCO} \\
        method{\scriptsize{\ph{+}}} & \scriptsize (img + text)\ph{.} & dataset & samples & \scriptsize R@1 & \scriptsize R@5 & \scriptsize R@10 & \scriptsize R@1 & \scriptsize R@5 & \scriptsize R@10 & \scriptsize R@1 & \scriptsize R@5 & \scriptsize R@10 & \scriptsize R@1 & \scriptsize R@5 & \scriptsize R@10 \\
        \shline
        \multicolumn{16}{c}{\scriptsize (a) comparisons with CLIP-\textbf{Base} baselines} \\
        \hline
        \scriptsize OpenAI CLIP-B/16\ph{+} & \scriptsize 86M + 63M & \scriptsize WIT-400M & \scriptsize 13B & \scriptsize 81.9 & \scriptsize 96.2 & \scriptsize 98.8 & \scriptsize 52.4 & \scriptsize 76.8 & \scriptsize 84.7 & \scriptsize 62.1 & \scriptsize 85.6 & \scriptsize 91.8 & \scriptsize 33.1 & \scriptsize 58.4 & \scriptsize 69.0 \\
        \rpink
        \scriptsize \eva-CLIP-B/16\phpink{+} & \scriptsize 86M + 63M & \scriptsize Merged-2B & \scriptsize \phpink{0}{8B} & \scriptsize \textbf{85.7} & \scriptsize \textbf{96.7} & \scriptsize \textbf{98.9} & \scriptsize \textbf{58.7} & \scriptsize \textbf{80.7} & \scriptsize \textbf{88.2} & \scriptsize \textbf{71.2} & \scriptsize \textbf{91.0} & \scriptsize \textbf{94.7} & \scriptsize \textbf{42.2} & \scriptsize \textbf{66.9} & \scriptsize \textbf{76.3} \\
        \scriptsize \demphs{OpenAI CLIP-L/14\ph{+}} & \demphs{\scriptsize 304M + 124M} & \demphs{\scriptsize WIT-400M} & \demphs{\scriptsize 13B} & \demphs{\scriptsize 85.2} & \demphs{\scriptsize 97.3} & \demphs{\scriptsize 99.0} & \demphs{\scriptsize 56.3} & \demphs{\scriptsize 79.3} & \demphs{\scriptsize 86.7} & \demphs{\scriptsize 65.2} & \demphs{\scriptsize 87.3} & \demphs{\scriptsize 92.0} & \demphs{\scriptsize 36.5} & \demphs{\scriptsize 61.0} & \demphs{\scriptsize 71.1} \\
        \shline
        \multicolumn{16}{c}{\scriptsize (b) comparisons with \textbf{larger} CLIP models} \\
        \hline
        \scriptsize OpenAI CLIP-L/14\ph{+} & \scriptsize 0.3B + 124M & \scriptsize WIT-400M & \scriptsize 13B & \scriptsize 85.2 & \scriptsize 97.3 & \scriptsize 99.0 & \scriptsize 56.3 & \scriptsize 79.3 & \scriptsize 86.7 & \scriptsize 65.2 & \scriptsize 87.3 & \scriptsize 92.0 & \scriptsize 36.5 & \scriptsize 61.0 & \scriptsize 71.1 \\
        \scriptsize OpenCLIP-L/14\ph{+} & \scriptsize 0.3B + 124M & \scriptsize LAION-2B & \scriptsize 32B & \scriptsize 88.7 & \scriptsize 98.4 & \scriptsize 99.2 & \scriptsize 62.1 & \scriptsize 83.4 & \scriptsize 90.3 & \scriptsize 75.0 & \scriptsize 92.5 & \scriptsize 95.6 & \scriptsize 46.1 & \scriptsize 70.7 & \scriptsize 79.4 \\
        \rpink
        \scriptsize \eva-CLIP-L/14\phpink{+} & \scriptsize 0.3B + 124M & \scriptsize Merged-2B & \scriptsize \phpink{0}{4B} & \scriptsize \textbf{89.7} & \scriptsize {98.6} & \scriptsize {99.2} & \scriptsize {63.7} & \scriptsize {84.3} & \scriptsize {90.4} & \scriptsize {77.3} & \scriptsize {93.6} & \scriptsize \textbf{96.8} & \scriptsize {47.5} & \scriptsize {71.2} & \scriptsize {79.7} \\
        \hline
        \scriptsize OpenAI CLIP-L/14+ & \scriptsize 0.3B + 124M & \scriptsize WIT-400M & \scriptsize 13B & \scriptsize 87.4 & \scriptsize 98.3 & \scriptsize 99.3 & \scriptsize 57.9 & \scriptsize 81.2 & \scriptsize 87.9 & \scriptsize 67.3 & \scriptsize 89.0 & \scriptsize 93.3 & \scriptsize 37.1 & \scriptsize 61.6 & \scriptsize 71.5 \\
        \rpink
        \scriptsize \eva-CLIP-L/14+ & \scriptsize 0.3B + 124M & \scriptsize Merged-2B & \scriptsize \phpink{0}6B & \scriptsize 89.2 & \scriptsize \textbf{98.9} & \scriptsize \textbf{99.6} & \scriptsize \textbf{64.1} & \scriptsize \textbf{85.2} & \scriptsize \textbf{90.8} & \scriptsize \textbf{77.9} & \scriptsize \textbf{94.2} & \scriptsize \textbf{96.8} & \scriptsize \textbf{47.9} & \scriptsize \textbf{71.7} & \scriptsize \textbf{80.0} \\
        \scriptsize \demphs{OpenCLIP-H/14\ph{+}} & \scriptsize \demphs{0.6B + 354M} & \scriptsize \demphs{LAION-2B} & \scriptsize \demphs{32B} & \scriptsize \demphs{90.8} & \scriptsize \demphs{99.3} & \scriptsize \demphs{99.7} & \scriptsize \demphs{66.0} & \scriptsize \demphs{86.1} & \scriptsize \demphs{91.9} & \scriptsize \demphs{77.8} & \scriptsize \demphs{94.1} & \scriptsize \demphs{96.6} & \scriptsize \demphs{49.5} & \scriptsize \demphs{73.4} & \scriptsize \demphs{81.5} 
    \end{tabular}
\vspace{-1.em}
\caption{\textbf{\eva-CLIP zero-shot retrieval performance.}}
\label{tab: clip zs retrieval}
\end{table*}

\paragraph{Zero-shot retrieval performance.}
\tblref{tab: clip zs retrieval} comprehensively reports the zero-shot image and text retrieval results on Flickr30K~\cite{flickr30K} and COCO~\cite{lin2014coco}. 
\eva-CLIP outperforms all the competitors with the same model size. 
While the zero-shot retrieval performance of \eva-CLIP is not as significant as classification compared to OpenCLIP-H, the results are still competitive. 
We speculate that the main reason for this difference is that retrieval tasks depend more on the capacity and capability of the language encoder compared to classification tasks.

\subsection{Object Detection and Segmentation}
\label{sec: od and seg}
In this section, we evaluate the transfer learning performance of \eva to mainstream object-level and pixel-level recognition benchmarks, namely, object detection and instance segmentation on COCO~\cite{lin2014coco} and LVIS~\cite{gupta2019lvis} in~\sref{sec: od}, as well as semantic segmentation on COCO-Stuff-164K~\cite{coco-stuff} and ADE20K~\cite{zhou2018ade} in~\sref{sec: seg}.

\begin{table}[!t]
\vspace{.5em}
\centering
\subfloat[
\textbf{Head-to-head comparisons} with the open-sourced ViTDet config.
\label{tab: base coco det baseline}
]{
\centering
\tablestyle{5pt}{1.2}
\begin{tabular}{y{51}|z{30}|x{22}x{22}|x{22}x{22}}
& enc. & \multicolumn{2}{c|}{{\scriptsize{COCO}} $\mathtt{val}$} & \multicolumn{2}{c}{{\scriptsize{LVIS}} $\mathtt{val}$} \\
method & \#params & \boxAP & \maskAP & \boxAP & \maskAP \\
\shline
\scriptsize ViTDet-B~\cite{li2022exploring} & 86M & 54.0 & 46.7 & 43.0 & 38.9 \\
\rpink
\scriptsize \eva-B & 86M & \textbf{55.5} & \textbf{47.1} & \textbf{47.1} & \textbf{41.4} \\
\end{tabular}
}
\vspace{-.4em}
\\
\subfloat[
\textbf{System comparisons} \textit{without} additional detection training data.
\label{tab: base coco det sota wo o365}
]{
\centering
\tablestyle{7.5pt}{1.2}
\begin{tabular}{y{51}|z{30}|x{44}x{44}}
& enc. & \multicolumn{2}{c}{{\scriptsize{COCO}} $\mathtt{val}$} \\
method & \#params & \boxAP & \maskAP \\
\shline
\scriptsize ViTDet-B~\cite{li2022exploring} & 86M & 56.0 & 48.0 \\
\scriptsize MViTv2-L~\cite{li2021improved} & 218M & 56.9 & 48.6 \\
\scriptsize MViTv2-H~\cite{li2021improved} & 667M & 57.1 & 48.8 \\
\rpink
\scriptsize \eva-B & \textbf{86M} & \textbf{58.9} & \textbf{50.7} \\
\end{tabular}
}
\vspace{-.5em}
\caption{\textbf{Object detection and instance segmentation results of \eva-B.}}
\label{tab: base od}
\vspace{-.5em}
\end{table}

\subsubsection{Object Detection and Instance Segmentation}
\label{sec: od}

For thoroughly evaluating the performance of \eva on object detection and instance segmentation tasks, we adopt the canonical Cascade Mask R-CNN~\cite{Mask-rcnn,cai2019cascade} as the task layer. 
This choice is motivated by its versatility in simultaneously performing both tasks, as well as its robustness and accuracy. 
To ensure fair comparisons with existing state-of-the-art methods, we essentially follow the training settings and architecture configurations of ViTDet~\cite{li2022exploring}, which includes large-scale jittering (LSJ) data augmentation~\cite{simple_copy_paste} and interleaved windowed and global attention mechanisms.

The model architecture as well as the hyper-parameters for COCO and LVIS are almost the same, except we use federated loss~\cite{zhou2021probabilistic} and repeat factor sampling~\cite{gupta2019lvis} following ViTDet on LVIS.
For LVIS, we use the IN-21K MIM pre-trained checkpoints of \eva for all experiments, as the COCO training images in the Merged-38M dataset include 10k images in LVIS \val set\footnote{In the Appendix, we show including unlabeled images from development / test set for MIM pre-training \textbf{does not} improve the final performance.}.

In the rest of this section, we evaluate \eva under three different transfer learning settings in~\tblref{tab: base od} and~\tblref{tab: large od}, including (i) a sanity check, (ii) a system-level comparison \textit{without} using additional detection data, and (iii) a system-level comparison \textit{with} additional intermediate detection fine-tuning.

\paragraph{(i) A sanity check.}
We first use the same open-sourced architectural configurations as ViTDet {\footnotesize \demphinline{(LSJ with 1024\suptext{2} crops, 4$\times$global attention blocks)}} to perform a head-to-head comparison.
In general, both \eva-B in~\tblref{tab: base coco det baseline} and \eva-L in~\tblref{tab: det head-to-head} can outperform the same- / larger-sized ViTDet \textit{w/} Cascade Mask R-CNN counterparts by a large margin, especially on LVIS.

\begin{table}[!t]
\vspace{-1.5em}
\centering
\subfloat[
\textbf{Head-to-head comparisons} with the open-sourced ViTDet config.
\label{tab: det head-to-head}
]{
\centering
\tablestyle{3.5pt}{1.2}
\begin{tabular}{y{65}|z{30}|x{24}x{24}|x{24}x{24}}
& enc. & \multicolumn{2}{c|}{{\scriptsize{COCO}} $\mathtt{val}$} & \multicolumn{2}{c}{{\scriptsize{LVIS}} $\mathtt{val}$} \\
method & \#params & \boxAP & \maskAP & \boxAP & \maskAP \\
\shline
\scriptsize ViTDet-L~\cite{li2022exploring} & 304M & 57.6 & 50.0 & 49.2 & 44.5 \\
\scriptsize ViTDet-H~\cite{li2022exploring} & 632M & 58.7 & \textbf{51.0} & 51.5 & 46.6 \\
\rpink
\scriptsize \eva-L & \textbf{304M} & \textbf{59.2} & 50.8 & \textbf{55.3} & \textbf{48.6} \\
\end{tabular}
}
\vspace{-.4em}
\\
\subfloat[
\textbf{System comparisons} \textit{without} additional detection training data.
\label{tab: det wo o365}
]{
\centering
\tablestyle{3.5pt}{1.2}
\begin{tabular}{y{65}|z{30}|x{24}x{24}|x{24}x{24}}
& enc. & \multicolumn{2}{c|}{{\scriptsize{COCO}} $\mathtt{val}$} & \multicolumn{2}{c}{{\scriptsize{LVIS}} $\mathtt{val}$} \\
method & \#params & \boxAP & \maskAP & \boxAP & \maskAP \\
\shline
\scriptsize ViTDet-L~\cite{li2022exploring} & 304M & 59.6 & 51.1 & 51.2 & 46.0 \\
\scriptsize ViTDet-H~\cite{li2022exploring} & 632M & 60.4 & 52.0 & 53.4 & 48.1 \\
\scriptsize RevCol-H~\cite{revcol} & 2158M & 61.1 & 53.0 & - & - \\
\rpink
\scriptsize \eva-L & \textbf{304M} & \textbf{62.3} & \textbf{53.8} & \textbf{60.1} & \textbf{53.5} \\
\end{tabular}
}
\vspace{-.4em}
\\
\subfloat[
\textbf{System comparisons} on COCO \textit{with} additional training on O365.
\label{tab: coco det w o365}
]{
\centering
\tablestyle{3.5pt}{1.2}
\begin{tabular}{y{65}|z{30}|x{24}x{24}|x{24}x{24}}
& enc. & \multicolumn{2}{c|}{{\scriptsize{COCO}} $\mathtt{val}$} & \multicolumn{2}{c}{{\scriptsize{COCO}} $\mathtt{test}$-$\mathtt{dev}$} \\
method & \#params & \boxAP & \maskAP & \boxAP & \maskAP \\
\shline
\scriptsize BEiT-3~\cite{beit3} & 1011M & -\suptext{\ph{tta}} & -\suptext{\ph{tta}} & 63.7\suptext{\demphs{tta}} & 54.8\suptext{\demphs{tta}} \\
\scriptsize FocalNet-H~\cite{yang2022focal} & 689M & 63.8\suptext{\ph{tta}} & -\suptext{\ph{tta}} & 63.9\suptext{\ph{tta}} & -\suptext{\ph{tta}} \\
\scriptsize FD-SwinV2-G~\cite{wei2022featdistill} & $\appp$3000M & -\suptext{\ph{tta}} & -\suptext{\ph{tta}} & 64.2\suptext{\demphs{tta}} & 55.4\suptext{\demphs{tta}} \\
\scriptsize InternImg-XL$\dag\dag$~\cite{internimage} & $\appp$600M & {64.2}\suptext{\demphs{tta}} & -\suptext{\ph{tta}} & {64.3}\suptext{\demphs{tta}} & -\suptext{\ph{tta}} \\
\scriptsize GDETRv2~\cite{chen2022group} & 632M & -\suptext{\ph{tta}} & -\suptext{\ph{tta}} & \textbf{64.5}\suptext{\demphs{tta}} & -\suptext{\ph{tta}} \\
\scriptsize \evaone~\cite{eva} & 1011M & \textbf{64.2}\suptext{\ph{tta}} & 55.0\suptext{\ph{tta}} & 64.4\suptext{\ph{tta}} & 55.5\suptext{\ph{tta}} \\
\rpink
\scriptsize \eva-L & \textbf{304M} & 64.1\suptext{\phpink{tta}} & \textbf{55.4}\suptext{\phpink{tta}} & \textbf{64.5}\suptext{\phpink{tta}} & \textbf{55.8}\suptext{\phpink{tta}} \\
\scriptsize \demphs{InternImg-H$\dag\dag$}~\cite{internimage} & \demphs{$\appp$2000M} & \demphs{65.0}\suptext{\demphs{tta}} & \demphs{-}\suptext{\ph{tta}} & \demphs{65.4}\suptext{\demphs{tta}} & \demphs{-}\suptext{\ph{tta}} \\
\end{tabular}
}
\vspace{-.4em}
\\
\subfloat[
\textbf{System comparisons} on LVIS \textit{with} additional training on O365.
\label{tab: lvis det w o365}
]{
\centering
\tablestyle{6.5pt}{1.2}
\begin{tabular}{y{60}|z{30}|x{44}x{44}}
& enc. & \multicolumn{2}{c}{{\scriptsize{LVIS}} $\mathtt{val}$} \\
method & \#params & \boxAP & \maskAP \\
\shline
\scriptsize \evaone~\cite{eva} & 1011M & 62.2\suptext{\ph{tta}} & 55.0 \\
\scriptsize InternImg-H$\dag\dag$~\cite{internimage} & $\appp$2000M & 63.2\suptext{\demphs{tta}} & - \\
\rpink
\scriptsize \eva-L & \textbf{304M} & \textbf{65.2}\suptext{\phpink{tta}} & \textbf{57.3} \\
\end{tabular}
}
\vspace{-.5em}
\caption{\textbf{Object detection and instance segmentation results of \eva-L.} \\ {\footnotesize \demphinline{``$\dag\dag$'': encoder parameters doubled using model composite technique~\cite{liang2022cbnet}}}}
\label{tab: large od}
\vspace{-1.em}
\end{table}

\paragraph{(ii) System comparisons \textit{w/o} additional detection data.}
In~\tblref{tab: base coco det sota wo o365} and~\tblref{tab: det wo o365}, we explore the limits of \textit{pure} MIM pre-trained \eva-B and -L representations in object detection and instance segmentation tasks.
To fully unleash the potential of \eva, we use an improved ViTDet configuration {\footnotesize \demphinline{(LSJ with 1536\suptext{2} crops, windowed attention with a size of 32, and 6$\times$ / 8$\times$global attention blocks for base- / large-sized models)}}.
Soft-NMS~\cite{bodla2017soft} is also applied.
For instance segmentation task, the classification score is calibrated~\cite{huang2019mask} via maskness~\cite{solo}.
The baselines we compared also adopt improved settings such as larger input resolution, Soft-NMS, \etc, and RevCol initializes HTC++~\cite{chen2019htc,liu2021swin} as the task layer, which is an improved version of Cascade Mask R-CNN we used.

Our experiments demonstrate that \eva significantly outperforms the same- and larger-sized counterparts, particularly on LVIS. 
These findings are consistent with our previous results in~\tblref{tab: base coco det baseline} and~\tblref{tab: det head-to-head}. 
We also encourage future work in representation learning to conduct more in-depth investigations on the \textit{original} pre-trained representations before adding more intermediate processes to chase the absolute performance.

\paragraph{(iii) System comparisons \textit{w/} additional O365 training.}
For the state-of-the-art detection system comparisons in~\tblref{tab: coco det w o365} and~\tblref{tab: lvis det w o365}, all methods use Object365 (O365)~\cite{o365} detection annotations for further performance improvements.
We additionally use EMA~\cite{ema} to update model weights. 
All results of \eva use single-scale evaluation while methods leveraging test-time augmentations are marked with ``tta'' superscript.
Methods that sacrifice instance segmentation ability in~\tblref{tab: coco det w o365} use the better-established DINO~\cite{detdino} as the detector.
Compared with other state-of-the-art approaches with much larger model sizes, our \eva is still quite competitive, especially on LVIS.

\begin{table}[!t] 
\vspace{-1.5em}
    \centering
    \tablestyle{4.5pt}{1.2}
    \begin{tabular}{l|z{30}x{30}x{40}|x{30}}
        & enc. & crop & extra & \scriptsize{ADE20K} \\
        method & \#params & size & labeled data & \mIoU \\
        \shline
        \multicolumn{5}{c}{\scriptsize{(a) comparisons with \textbf{based}-sized encoders}} \\
        \hline 
        \scriptsize BEiTv2-B~\cite{beitv2} & 86M & 512\suptext{2} & \xmark & 53.1  \\
        \scriptsize BEiTv2-B~\cite{beitv2} & 86M & 512\suptext{2} & \scriptsize IN-21K & 53.5 \\
        \rpink 
        \scriptsize \eva-B & \textbf{86M} & 512\suptext{2} & \xmark & \textbf{55.3} \\ 
        \scriptsize \demphs{DeiT-III-L}~\cite{deit3} & \demphs{304M} & \demphs{512\suptext{2}} & \demphs{\scriptsize IN-21K} & \demphs{54.6} \\
        \scriptsize \demphs{InternImage-XL}~\cite{internimage} & \demphs{335M} & \demphs{640\suptext{2}} & \demphs{\scriptsize IN-21K} & \demphs{55.0} \\
        \scriptsize \demphs{ConvNeXt V2-H}~\cite{woo2023convnext} & \demphs{707M} & \demphs{512\suptext{2}} & \demphs{\xmark} & \demphs{55.0} \\
        \shline
        \multicolumn{5}{c}{\scriptsize{(b) comparisons with \textbf{larger}-sized encoders}} \\
        \hline
        \scriptsize BEiTv2-L~\cite{beitv2} & 304M & 512\suptext{2} & \xmark & 56.7 \\
        \scriptsize BEiTv2-L~\cite{beitv2} & 304M & 512\suptext{2} & \scriptsize IN-21K & 57.5 \\
        \rpink
        \scriptsize \eva-L & \textbf{304M} & 512\suptext{2} & \xmark & \textbf{59.8} \\
        \rpink
        \scriptsize \eva-L+ & \textbf{304M} & 640\suptext{2} & \xmark & \textbf{60.1} \\ 
        \scriptsize \demphs{ConvNeXt V2-H}~\cite{woo2023convnext} & \demphs{707M} & \demphs{640\suptext{2}} & \demphs{\scriptsize IN-21K} & \demphs{57.0} \\
        \scriptsize \demphs{RevCol-H}~\cite{revcol} & \demphs{2158M} & \demphs{640\suptext{2}} & \demphs{168M} & \demphs{57.8} \\        
        \scriptsize \demphs{SwinV2-G}~\cite{swinv2} & \demphs{$\appp$3000M} & \demphs{896\suptext{2}} & \demphs{70M} & \demphs{59.3} \\
        \scriptsize \demphs{InternImage-H}~\cite{internimage} & \demphs{1080M} & \demphs{896\suptext{2}} & \demphs{\scriptsize IN-21K} & \demphs{59.9} \\
        \end{tabular}
        \vspace{-1.em}
        \caption{\textbf{Semantic segmentation with UperNet.} All methods use single-scale evaluation. \\ {\footnotesize \demphinline{``+'': using larger input resolution and segmentation head dimension}}}
        \label{tab: upernet sem seg}
\end{table}

\begin{table}[!t] 
    \centering
    \tablestyle{3.6pt}{1.2}
    \begin{tabular}{y{66.6}|rc|c|cc}
        & enc. & crop & \scriptsize{COCO164K} & \multicolumn{2}{c}{\scriptsize{ADE20K}} \\
        method & \#params & size & \mIoUss & \mIoUss & \mIoUms \\
        \shline
        \scriptsize RevCol-H & 2158M & 640\suptext{2} & - & 60.4 & 61.0 \\
        \scriptsize BEiTv2-L w/ ViT-Ada. & 304M & 896\suptext{2} & 52.3 & 61.2 & 61.5 \\
        \scriptsize \evaone w/ ViT-Ada. & 1011M & 896\suptext{2} & 53.4 & 61.5 & \textbf{62.3} \\ 
        \rpink
        \scriptsize \eva-L & \textbf{304M} & 640\suptext{2} & \textbf{53.7} & \textbf{61.7} & 62.0 \\
        \end{tabular}
        \vspace{-1.em}
        \caption{\textbf{Semantic segmentation with Mask2Former.} ``\mIoUss / \mIoUms'': mIoU using single-scale / multi-scale evaluation.}
        \label{tab: m2f sem seg}
        \vspace{-.5em}
\end{table}

\subsubsection{Semantic Segmentation}
\label{sec: seg}

We comprehensively evaluate the semantic segmentation performance of \eva-B and -L models using two different task layers: UperNet~\cite{xiao2018upernet} and Mask2Former~\cite{mask2former} on two widely adopted benchmarks: ADE20K~\cite{zhou2018ade} and COCO-Stuff-164K~\cite{coco-stuff}.
Notably, unlike previous mainstream approaches that involve additional fine-tuning such as using IN-21K intermediate fine-tuned models for semantic segmentation, we primarily evaluate \textit{pure} MIM pre-trained representations of \eva.

\paragraph{UperNet results.}
As shown in~\tblref{tab: upernet sem seg}, both pure MIM pre-trained \eva-B and -L models with UperNet segmenter significantly outperform the same-sized BEiTv2 models without or \textit{with} the additional 90-epoch IN-21K intermediate fine-tuning.
Furthermore, our representation can outperform larger pre-trained counterparts such as ConvNeXt V2, InternImage, \etc, and achieves up to 60.1 \mIoU with single-scale evaluation.

\paragraph{Mask2Former results.}
\tblref{tab: m2f sem seg} shows the state-of-the-art model comparisons on COCO-Stuff-164K and ADE20K benchmarks.
Models for ADE20K segmentation are initialized from COCO-Stuff-164K pre-trained representations as long as the COCO-Stuff-164K results are reported.
BEiTv2-L and \evaone also utilize ViT-Adapter {\footnotesize \demphinline{(ViT-Ada. in~\tblref{tab: m2f sem seg})}}~\cite{vitadapt} for architectural improvements.

Compared with larger models using the Mask2Former task layer, our approach is still quite performant, and creates new state-of-the-art results with large-sized models on both COCO-Stuff-164K and ADE20K semantic segmentation benchmarks.

\subsection{Summary of All Evaluations}
In~\sref{sec: experiments}, we demonstrate the excellent transfer learning ability of pre-trained \eva representations on a large diversity of downstream tasks.
Although all tasks / benchmarks we evaluated are at the core of computer vision, here we would like to (re-)emphasize the importance of the ones related to \eva-CLIP: 
not only for the promising zero-shot transferability, but also because the vision features from \eva-CLIP are \textit{well aligned} with natural language that comes with much broader supervision than pure vision signals / features as well as fixed set of pre-determined label sets.
Therefore, we hope \eva-CLIP can serve as a basic building blocks and provide more robust vision features for future multi-modal systems.

\section{Related Work}
Some previous advancements in representation learning do not necessarily come with entirely new ideas or novel approaches.
The GPT series~\cite{radford2018improving,radford2019language,gpt3,ouyang2022training} achieve quantitative changes that transform the landscape of scientific research by continuously scaling the simplest language modeling.
RoBERTa~\cite{liu2019roberta} present a detailed replication study of BERT pre-training~\cite{devlin2018bert} that carefully measures the impact of many key hyper-parameters, training data and objectives, which results in greatly improved bidirectional language representations.
DeiT~\cite{touvron2021training} and RSB~\cite{rsb} closely evaluate the training recipe for smaller-sized plain ViTs~\cite{dosovitskiy2020vit} and ResNets~\cite{resnet} respectively, while ConvNeXt~\cite{convnext} collectively examines previous architectural advancements for the next-gen ConvNets model design.
\cite{beyer2022knowledge} empirically shows that a robust and effective recipe of knowledge distillation makes state-of-the-art large-scale image classification models affordable in practice.

Inspired by the spirits of these works, this paper provides a thorough evaluation of MIM visual representation learning~\cite{bao2021beit,zhou2021ibot,xie2021simmim,he2021masked} that significantly bridge the gap between large-scale visual representations that achieve state-of-the-art performance and models that are affordable and accessible for the wider research community.

\section{Discussion and Conclusion}
In this work, we aim to contribute to the ongoing research on visual and vision-language representation learning.
Instead of proposing an entirely new architecture or method, we present an in-depth evaluation of the existing MIM pre-training with CLIP vision features as the pretext task's targets.
Our experiments demonstrated that if robustly optimized, this approach is capable of producing highly performant, affordable, and transferable representations that outperform larger state-of-the-art specialized models.

Our analysis has revealed that base- \& large-sized \eva models can be effectively leveraged to obtain compact and expressive CLIP representations, which have the potential to facilitate modular, reusable, and scalable model design in the future~\cite{dalle2,alayrac2022flamingo,chen2022pali,blip2}. 
Our findings on moderate-sized models can also serve as a valuable reference for future research on model and representation scaling.

\begin{figure}[t!]
    \centering
    \includegraphics[width=\linewidth]{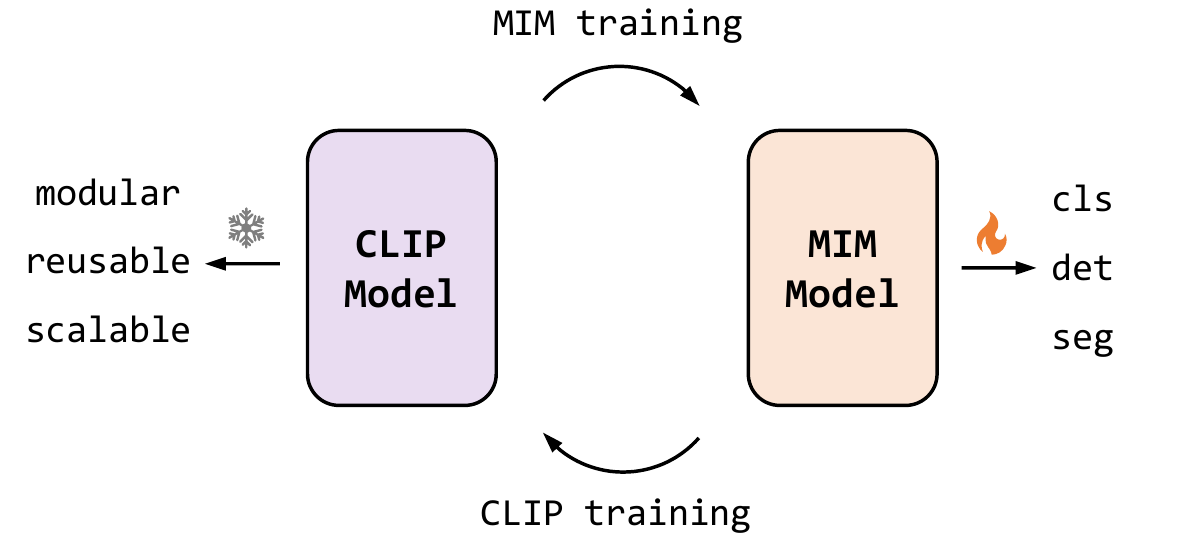}
    \vspace{-1.8em}
    \caption{\textbf{Alternate learning of MIM and CLIP representations.} Starting with a off-the-shelf CLIP{\footnotesize \demphinline{(\eg, OpenAI CLIP~\cite{clip})}}, alternate training of the pure MIM visual representations as well as vision-language CLIP representations can improve both MIM and CLIP performances in a bootstrapped manner. The MIM representations can be used to fine-tune various downstream tasks while the (frozen) CLIP representations enable modular, reusable and scalable next-gen model design.}
    \vspace{-.8em}
    \label{fig: conclusion}
\end{figure}

Furthermore, in combination with \evaone~\cite{eva}, we demonstrate that alternate training of the pure MIM visual representations as well as vision-language CLIP representations can improve both MIM and CLIP performances in a bootstrapped manner (\figref{fig: conclusion}).
This suggests a promising and scalable approach for pre-training both vision and vision-language representations of various sizes, which warrants further exploration in future research.

\section*{Acknowledgement}
\label{sec: ack}
We would like to thank Hanxiao Qu, Yan Tian, Yemin Shi and Xigang Cao for their help on GPU resources.
Zhao Xue, Quanyue Ma and Bowen Zhang for their help on datasets and benchmarks, and other colleagues at Beijing Academy of Artificial Intelligence for support throughout this project.
We thank Wen Wang for constructive discussions on object detection \& instance segmentation tasks, and Qiang Chen for constructive discussions on model weight initialization.


\appendix

\begin{table}[!t]
    \centering
    \tablestyle{5pt}{1.2}
    \begin{tabular}{l|cx{27}x{25}x{25}|x{25}}
        & \scriptsize MIM & {\scriptsize{IN-1K}} ft & & {\scriptsize{IN-21K}} & {\scriptsize{IN-1K}} \\ 
        method & teacher & img size & \scriptsize FLOPs & label? & top-1 \\
        \shline
        \multicolumn{6}{c}{\scriptsize{(a) ViT-Base model (86M), IN-1K ft number of tokens = 196}} \\
        \hline
        \scriptsize BEiTv2-B~\cite{beitv2} & \scriptsize VQKD-B & 224\suptext{2} & 18G & \xmark & 85.5 \\
        \scriptsize dBOT-B~\cite{dBOT} & \scriptsize CLIP-B & 224\suptext{2} & 18G & \xmark & 85.7 \\
        \scriptsize \demphs{BEiTv2-B}~\cite{beitv2} & \scriptsize VQKD-B & \demphs{224\suptext{2}} & \demphs{18G} & \demphs{\cmark} & \demphs{86.5} \\
        \rpink
        \scriptsize \eva-B & \scriptsize \evaone-CLIP & 196\suptext{2} & 18G & \xmark & \textbf{87.0} \\
        \shline
        \multicolumn{6}{c}{\scriptsize{(b) ViT-Large model (304M), IN-1K ft number of tokens = 196}} \\
        \hline
        \scriptsize BEiTv2-L~\cite{beitv2} & \scriptsize VQKD-B & 224\suptext{2} & 62G & \xmark & 87.3 \\
        \scriptsize dBOT-L~\cite{dBOT} & \scriptsize CLIP-L & 224\suptext{2} & 62G & \xmark & 87.8 \\
        \scriptsize \demphs{BEiTv2-L}~\cite{beitv2} & \demphs{\scriptsize VQKD-B} & \demphs{224\suptext{2}} & \demphs{62G} & \demphs{\cmark} & \demphs{88.4} \\
        \rpink
        \scriptsize \eva-L & \scriptsize \evaone-CLIP & 196\suptext{2} & 62G & \xmark & \textbf{88.9} \\
    \end{tabular}
    \vspace{-.5em}
    \caption{\textbf{Head-to-head comparisons of based- and large-sized models on IN-1K val set classification.} The fine-tuning settings are relatively moderate with the same compute budget for each model.}
    \label{tab: b&l 1k cls}
\end{table}

\section{Appendix}
\label{app}

\subsection{Architecture}
\label{app: arch}

\paragraph{SwiGLU FFN.}
The position-wise feedforward network (FFN) in the original ViT design~\cite{dosovitskiy2020vit} is a multi-layer perceptron (MLP) contains two layers (represented by the weight matrices $W_1$ and $W_2$, biases are omitted) with a GELU~\cite{hendrycks2016gaussian} activation function, denoted as $\operatorname{FFN}_{\operatorname{MLP}}$.
Formally,
\begin{flalign}
\ \operatorname{FFN}_{\operatorname{MLP}}\left(x, W_1, W_2\right)=\operatorname{GELU}\left(x W_1\right) W_2. &&
\end{flalign}
SwiGLU FFN~\cite{shazeer2020glu} replace the first transformation of the original ViT's FFN with a variant of the Gated Linear Unit (GLU)~\cite{dauphin2017language} with a SiLU {\footnotesize \demphinline{($\operatorname{SiLU} = x * \operatorname{sigmoid}(x)$)}} activation function~\cite{hendrycks2016gaussian,ramachandran2017searching}, Formally,
\begin{flalign}
\ \operatorname{FFN}_{\operatorname{SwiGLU}}\left(x, U, V, W\right)=\left(\operatorname{SiLU}(x U) \odot x V\right) W, &&
\end{flalign}
where $\odot$ is the element-wise product. 

To keep the number of parameters and the amount of computation constant, we reduce the hidden units (the output dimension of $U$ and $V$ and the input dimension of $W$) of $\operatorname{FFN}_{\operatorname{SwiGLU}}$ by a factor of 2/3 when comparing these layers to the original $\operatorname{FFN}_{\operatorname{MLP}}$.

\paragraph{Normalization.}
We use sub-LN~\cite{wang2022foundation} {\footnotesize \demphinline{(We find the inner attention LN unnecessary so we drop it)}} as the default normalization scheme for \eva-B and -L blocks.
For the tiny- and small-sized model, we find using the default pre-LN configuration following~\cite{dosovitskiy2020vit,bao2021beit} is sufficient.

\paragraph{RoPE} is a type of position embedding that unifies absolute as well as relative potential representations, and is widely adopted in state-of-the-art language models~\cite{black2022gpt,chowdhery2022palm,chen2022pali}.
For a detailed description of RoPE, please refer to~\cite{su2021roformer,ropeeleutherai}.
Our implementation is based on the open-sourced~\cite{ropetorch}.

In brief, RoPE twists / rotates the input embedding {\footnotesize \demphinline{(without changing the norm)}} such that the attention of a token at position $m$ to a token at position $n$ is linearly dependent on $m - n$.
Notably, unlike the conventional relative position representations that inject the positional information into the attention matrix, RoPE only manipulates $q$, $k$ vectors.
So RoPE is naturally compatible with off-the-shelf fused high-performance MHSA operators such as~\cite{dao2022flashattention,xFormers2022}.

\paragraph{RoPE} is a type of position embedding that unifies absolute as well as relative potential representations, and is widely adopted in state-of-the-art language models~\cite{black2022gpt,chowdhery2022palm,chen2022pali}.
For a detailed description of RoPE, please refer to~\cite{su2021roformer,ropeeleutherai}.
Our implementation is based on the open-sourced~\cite{ropetorch}.

In brief, RoPE twists / rotates the input embedding {\footnotesize \demphinline{(without changing the norm)}} such that the attention of a token at position $m$ to a token at position $n$ is linearly dependent on $m - n$.
Notably, unlike the conventional relative position representations that inject the positional information into the attention matrix, RoPE only manipulates $q$, $k$ vectors.
So RoPE is naturally compatible with off-the-shelf fused high-performance MHSA operators such as~\cite{dao2022flashattention,xFormers2022}.

\paragraph{Weight Initialization.}
We use $\mathtt{xavier}$ $\mathtt{normal}$~\cite{xnorm} to initialize all weights in \trv blocks.
The weight matrices in MHSA and FFN are sampled from $\mathcal{N} \sim (0, \text{std}^2)$, where std is $\sqrt{ 2 / (\text{dim}_{\text{in}} + \text{dim}_{\text{out}})}$.

\begin{table}[!t]
    \centering
    \tablestyle{4pt}{1.2}
    \begin{tabular}{x{40}|x{40}|x{25}x{25}|x{25}x{25}}
        & & \multicolumn{2}{c|}{{\scriptsize{IN-1K}} top-1} &  \multicolumn{2}{c}{\demphs{{\scriptsize{IN-V2}} top-1}} \\ 
        method & optimizer & $\mathtt{fp16}$ & $\mathtt{bf16}$ & \demphs{$\mathtt{fp16}$} & \demphs{$\mathtt{bf16}$} \\
        \shline
        & SGD & 88.40 & 88.37 & \demphs{79.73} & \demphs{79.67} \\
        \scriptsize \eva-B & AdamW & \evadefault{88.57} & \textbf{88.58} & \evadefault{\demphs{79.78}} & \demphs{79.74} \\
        & Lion & 88.52 & 88.50 & \demphs{\textbf{79.97}} & \demphs{79.96} \\
        \hline
        & SGD & 89.87 & 89.84 & \demphs{82.15} & \demphs{82.17} \\
        \scriptsize \eva-L & AdamW & \evadefault{89.98} & 89.95 & \evadefault{\demphs{82.43}} & \demphs{\textbf{82.61}} \\
        & Lion & 89.97 & \textbf{90.00} & \demphs{82.19} & \demphs{82.37} \\
    \end{tabular}
    \vspace{-.5em}
    \caption{\textbf{Study of different numerical precisions and optimizers on IN-1K classification fine-tuning.} To explore the limit of \eva representation, all pre-trained models are fine-tuned at a resolution of 448\suptext{2} with IN-21K intermediate fine-tuning following the most performant settings in~\tblref{tab: more scaling, better results}.}
    \label{tab: optim & precision}
\end{table}

\begin{table}[!t]
    \centering
    \tablestyle{4.2pt}{1.2}
    \begin{tabular}{l|c|r|x{50}x{50}}
        & & enc. & \multicolumn{2}{c}{best {\scriptsize{IN-1K}} top-1} \\ 
        method & role & \#params & \scriptsize \textit{w/o} IN-21K ft & \scriptsize \textit{w/} IN-21K ft \\
        \shline
        \scriptsize \evaone-CLIP$\dag$ & teacher & 1011M & 89.4 & 89.5 \\
        \scriptsize \eva-L & student & 304M & 89.6 & 90.0 \\
    \end{tabular}
    \vspace{-.5em}
    \caption{\textbf{Indigo blue comes from indigo.} With sufficient pre-training, \eva-L with 304M-parameter is able to surpass its teacher with 1011M-parameter in IN-1K image classification. \\ {\footnotesize \demphinline{``$\dag$'': fine-tuned CLIP vision encoder}}}
    \label{tab: student beats teacher}
\end{table}

\subsection{Additional Results for Image Classification}
\label{app: additional results cls}

\paragraph{\eva-B and -L.}
In~\tblref{tab: b&l 1k cls}, we show that sufficiently pre-trained pure MIM \eva representations {\footnotesize \demphinline{(\textit{w/o} IN-21K intermediate fine-tuning)}} outperform some previous leading approaches {\footnotesize \demphinline{(even \textit{w/} intermediate fine-tuning)}}.

\paragraph{Precisions and optimizers.}
In~\tblref{tab: optim & precision}, we show that sufficiently pre-trained \eva representations are robust enough that can be fine-tuned using various numerical precisions {\footnotesize \demphinline{(\eg, $\mathtt{fp16}$ and $\mathtt{bf16}$)}} and optimizers {\footnotesize \demphinline{(\eg, Lion~\cite{lion}, AdamW~\cite{adam,Loshchilov2019adamw}, and SGD~\cite{sgd})}}. 
Remarkably, the fine-tuning can be done using the SGD optimizer with only little performance drop.

\paragraph{The student is the master.}
\tblref{tab: student beats teacher} distinguishes MIM from conventional knowledge distillation~\cite{hinton2015distilling} in the context of ``pre-training \& fine-tuning'' paradigm.

\subsection{Data Contamination in MIM Pre-training: \\ A Case Study}
\label{app: data contamination}

We provide a case study about the impact of data contamination in MIM pre-training when transferred to object detection and instance segmentation tasks.
In short, we find the impact is minor.

We pre-train two \eva-L models, one uses the Merged-38M unlabeled images for MIM pre-training, and the other 
uses the images from IN-21K as the pre-training data.
Both models are pre-trained with 1M steps with a batch size of 2k.
Other settings \& configurations are the same.
Notice that the Merged-38M unlabeled images contain all Object365 (O365)~\cite{o365} \tset set images, as well as 15k out of 20k LVIS~\cite{gupta2019lvis} \val set images {\footnotesize \demphinline{(the Merged-38M images contain all the COCO training images, and LVISv1.0 \val split also contains 15k images from the COCO training set)}}.

\begin{table}[!t]
    \centering
    \tablestyle{5.3pt}{1.2}
    \begin{tabular}{x{40}|x{40}|x{20}x{20}|x{25}x{25}}
        & {\scriptsize MIM to O365} & \multicolumn{2}{c|}{\scriptsize MIM to LVIS} & \multicolumn{2}{c}{\scriptsize MIM to O365 to LVIS} \\
        & & \multicolumn{2}{c|}{\scriptsize (\tblref{tab: det head-to-head})} & \multicolumn{2}{c}{\scriptsize (\tblref{tab: lvis det w o365})} \\
        {\scriptsize MIM} pt data & \scriptsize \boxAP & \scriptsize \boxAP & \scriptsize \maskAP & \scriptsize \boxAP & \scriptsize \maskAP \\
        \shline
        \scriptsize Merged-38M & \scriptsize 50.57 & \scriptsize 55.34 & \scriptsize 48.74 & \scriptsize 65.42 & \scriptsize 57.42 \\
        \rpink
        \scriptsize IN-21K & \scriptsize 50.47 & \scriptsize 55.28 & \scriptsize 48.59 & \scriptsize 65.22 & \scriptsize 57.32 \\
    \end{tabular}
    \vspace{-.5em}
    \caption{\textbf{The impact of data contamination in MIM pre-training} when transferred to object detection \& instance segmentation tasks. The setting in \colorbox{02pink!20}{pink} is the default setting we used in~\tblref{tab: large od} for LVIS \val set evaluation.}
    \label{tab: data contamination}
\end{table}

\begin{table}[!t]
    \centering
    \tablestyle{4.pt}{1.2}
    \scriptsize
    \begin{tabular}{c|x{35}x{35}|x{35}x{35}}
        & \multicolumn{2}{c|}{based-sized model (86M)} & \multicolumn{2}{c}{larged-sized model (304M)} \\
        & ViT & \trv & ViT & \trv \\
        \shline
        throughput (img / s) & 1600 & 2226 & 554 & 636 \\
    \end{tabular}
    \vspace{-.5em}
    \caption{\textbf{Inference throughput comparisons using one A100 GPU.} The batch size is 1024. The number of patch tokens is 196. The architecture of ViT follows BEiT series~\cite{bao2021beit,beitv2} {\footnotesize \demphinline{(with rel. PE~\cite{rpe} and LayerScale~\cite{cait})}}.}
    \label{tab: throughput}
\end{table}

We study the transfer learning performance in three different settings: 

(i) Directly transfer pure MIM pre-trained \eva representations to O365 {\footnotesize \demphinline{(MIM to O365)}}, The performance is evaluated using the O365 \tset set\footnote{The \tset set images are publicly available. We have permission to access the annotations.}{\footnotesize \demphinline{(the \tset set is a very large and challenging benchmark with \app200k images and \app2.5M instances in 365 different categories)}}.

(ii) Directly transfer pure MIM pre-trained \eva representations to LVIS {\footnotesize \demphinline{(MIM to LVIS, \tblref{tab: det head-to-head})}}.
The performance is evaluated using LVIS \val set {\footnotesize \demphinline{(the \val set is a long-tail, large-vocabulary challenging benchmark with \app20k images and \app0.25M federated annotated instances in more than 1.2k different categories)}}.

(iii) Transfer the \eva representations with additional O365 intermediate fine-tuning to LVIS {\footnotesize \demphinline{(MIM to O365 to LVIS, \tblref{tab: lvis det w o365})}}.
The performance is evaluated using LVIS \val set.

The results are summarized in~\tblref{tab: data contamination}.
Overall, we find including unlabeled images from the development / test set for MIM pre-training has little impact on the final performance.

These experiments are motivated by our initial use of the Merged-38M pre-trained representation for LVIS \val set evaluation, which resulted in unintended use of unlabeled images from the development / test set for MIM pre-training, similar to the issue raised in \cite{longmae}.
\cite{cliph} also reports a small percentage of images from IN-1K along with its variants, Flickr30K and COCO were detected in the LAION-400M dataset.
This data contamination issue raises concerns about the validity of downstream benchmarks when a large number of unlabeled images are used for pre-training.
While it is possible to identify and remove all duplicates for existing benchmarks, it may be infeasible to do so on already pre-trained models for future benchmarks or in real-world applications.
Nonetheless, we believe that this issue should not hinder progress in data scaling for future representation learning studies.

\subsection{Implementation Details}

In this section, we summarize the training / evaluation settings, configurations, and hyper-parameters.

\subsubsection{MIM pre-training} 
\paragraph{\eva MIM pre-training setting.}
See~\tblref{tab: pt cfg}.

\subsubsection{Image Classification}

\paragraph{\trv throughput.}
See~\tblref{tab: throughput}.

\paragraph{Intermediate fine-tuning setting for IN-21K.} 
See~\tblref{tab: 21k ft cfg}.

\paragraph{Fine-tuning setting for IN-1K (\textit{w/} IN-21K intermediate fine-tuning).}
See~\tblref{tab: 21k to 1k ft cfg}.

\paragraph{Fine-tuning setting for IN-1K (\textit{w/o} IN-21K intermediate fine-tuning).}
See~\tblref{tab: 1k ft cfg}.

\subsubsection{Contrastive Language-Image Pre-training}

\paragraph{\eva enhanced CLIP training setting.} 
See~\tblref{tab: clip cfg}.

\subsubsection{Object Detection and Instance Segmentation}

\paragraph{O365 intermediate fine-tuning.}
See~\tblref{tab: o365 cfg}.

\paragraph{COCO head-to-head comparisons.}
See~\tblref{tab: coco head to head cfg}.

\paragraph{LVIS head-to-head comparisons.}
See~\tblref{tab: lvis head to head cfg}.

\paragraph{COCO system-level comparisons (\textit{w/o} O365 intermediate fine-tuning).}
See~\tblref{tab: coco wo o365 sota cfg}.

\paragraph{LVIS system-level comparisons (\textit{w/o} O365 intermediate fine-tuning).}
See~\tblref{tab: lvis wo o365 sota cfg}.

\paragraph{COCO system-level comparisons (\textit{w/} O365 intermediate fine-tuning).}
See~\tblref{tab: coco w o365 sota cfg}.

\paragraph{LVIS system-level comparisons (\textit{w/} O365 intermediate fine-tuning).}
See~\tblref{tab: lvis w o365 sota cfg}.

\subsubsection{Semantic Segmentation}

\paragraph{Using UperNet on ADE20K.}
See~\tblref{tab: seg upernet ade cfg}.

\paragraph{Using Mask2Former on COCO-Stuff-164K.}
See~\tblref{tab: seg m2f coco cfg}.

\paragraph{Using Mask2Former on ADE20K.}
See~\tblref{tab: seg m2f ade cfg}.

\begin{table}[h!]
\vspace{7.5em}
\centering
\tablestyle{7.5pt}{1.2}
\scriptsize
\begin{tabular}{l|c}
config & \eva-Ti / -S / -B / -L \\
\shline
enc. weight initialization & $\mathtt{xavier}$ $\mathtt{normal}$ random initialization~\cite{xnorm} \\
MIM teacher & \evaone-CLIP vision encoder~\cite{eva} \\

image data source & IN-21K / IN-21K / IN-21K / Merged-38M \\

peak learning rate & 3e-3 / 3e-3 / 1.5e-3 / 1.5e-3 \\
learning rate schedule & cosine decay \\

optimizer & AdamW~\cite{adam,Loshchilov2019adamw} \\
optimizer hyper-parameters & $\beta_1$, $\beta_2$, $\epsilon$ = 0.9, 0.98, 1e-6 \\
weight decay & 0.05 \\

input resolution & 224\suptext{2} \\
patch size & 14\suptext{2} \\
masking ratio & 40\% \\

batch size & 4k / 4k / 2k / 2k \\
training steps & 0.85M / 0.85M / 1M / 1M \\
training epochs & 240 / 240 / 150 / 56 \\
warmup epochs & 1 \\

drop path~\cite{huang2016deep} & 0.0 / 0.0 / 0.0 / 0.1 \\
random resized crop & (0.2, 1) \\

numerical precision & $\mathtt{DeepSpeed}$ $\mathtt{fp16}$~\cite{rasley2020deepspeed} \\
ZeRO optimizer~\cite{ramesh2021zero} & stage 0 or 1 \\

\end{tabular}
\vspace{-.5em}
\caption{MIM pre-training setting.}
\label{tab: pt cfg}
\end{table}

\begin{table}[h!]
\centering
\tablestyle{8pt}{1.2}
\scriptsize
\begin{tabular}{l|c}
config & \eva-B / -L \\
\shline

enc. weight initialization & MIM pre-trained \eva (\tblref{tab: pt cfg}) \\

peak learning rate & 3e-4 \\
layer-wise lr decay~\cite{clark2020electra, bao2021beit} & 0.70 / 0.75 \\
learning rate schedule & cosine decay \\

optimizer & AdamW~\cite{adam,Loshchilov2019adamw} \\
optimizer hyper-parameters & $\beta_1$, $\beta_2$, $\epsilon$ = 0.9, 0.999, 1e-8 \\
weight decay & 0.05 \\

input resolution & 448\suptext{2} \\
patch size & 14\suptext{2} \\
batch size & 2048 \\
training epochs & 40 / 30 \\
warmup epochs & 1 \\

drop path~\cite{huang2016deep} & 0.10 / 0.15 \\
label smoothing~\cite{szegedy2016rethinking} & 0.1 \\

augmentation & RandAug (9, 0.5)~\cite{cubuk2020randaugment} \\
random resized crop & (0.2, 1) \\

numerical precision & $\mathtt{DeepSpeed}$ $\mathtt{fp16}$~\cite{rasley2020deepspeed} \\
ZeRO optimizer~\cite{ramesh2021zero} & stage 0 or 1 \\

\demphs{ema~\cite{ema}} & \demphs{\xmark} \\
\demphs{cutmix~\cite{yun2019cutmix}} & \demphs{\xmark} \\
\demphs{mixup~\cite{zhang2017mixup}} & \demphs{\xmark} \\
\demphs{random erasing~\cite{zhong2020random}} & \demphs{\xmark} \\

\end{tabular}
\vspace{-.5em}
\caption{Intermediate fine-tuning setting for \textbf{IN-21K}.}
\label{tab: 21k ft cfg}
\end{table}

\begin{table}[b!]
\centering
\tablestyle{8pt}{1.2}
\scriptsize
\begin{tabular}{l|c}
config & \eva-B / -L \\
\shline

enc. weight initialization & IN-21K fine-tuned \eva (\tblref{tab: 21k ft cfg}) \\

peak learning rate & 5e-5 / 2e-5 \\
layer-wise lr decay~\cite{clark2020electra, bao2021beit} & 0.80 / 0.85 \\
learning rate schedule & cosine decay \\

optimizer & AdamW~\cite{adam,Loshchilov2019adamw} \\
optimizer hyper-parameters & $\beta_1$, $\beta_2$, $\epsilon$ = 0.9, 0.999, 1e-8 \\
weight decay & 0.05 \\

input resolution & 448\suptext{2} \\
patch size & 14\suptext{2} \\
batch size & 512 \\
training epochs & 15 / 20 \\
warmup epochs & 2 \\

drop path~\cite{huang2016deep} & 0.15 \\
label smoothing~\cite{szegedy2016rethinking} & 0.2 \\

augmentation & RandAug (9, 0.5)~\cite{cubuk2020randaugment} \\
random resized crop & (0.08, 1) \\
test crop ratio & 1.0 \\

numerical precision & $\mathtt{DeepSpeed}$ $\mathtt{fp16}$~\cite{rasley2020deepspeed} \\
ZeRO optimizer~\cite{ramesh2021zero} & stage 0 or 1 \\

ema~\cite{ema} & 0.9999 \\
\demphs{cutmix~\cite{yun2019cutmix}} & \demphs{\xmark} \\
\demphs{mixup~\cite{zhang2017mixup}} & \demphs{\xmark} \\
\demphs{random erasing~\cite{zhong2020random}} & \demphs{\xmark} \\

\end{tabular}
\vspace{-.5em}
\caption{Fine-tuning setting for \textbf{IN-1K} (\textbf{\textit{w/}} \textbf{IN-21K} intermediate fine-tuning).}
\label{tab: 21k to 1k ft cfg}
\end{table}

\begin{table}[t!]
\centering
\tablestyle{8pt}{1.2}
\scriptsize
\begin{tabular}{l|c}
config & \eva-Ti / -S / -B / -L \\
\shline

enc. weight initialization & MIM pre-trained \eva (\tblref{tab: pt cfg}) \\

peak learning rate & 2e-4 / 1e-4 / 1e-4 / 7e-5  \\
layer-wise lr decay~\cite{clark2020electra, bao2021beit} & 0.90 / 0.80 / 0.70 / 0.80 \\
learning rate schedule & cosine decay \\

optimizer & AdamW~\cite{adam,Loshchilov2019adamw} \\
optimizer hyper-parameters & $\beta_1$, $\beta_2$, $\epsilon$ = 0.9, 0.999, 1e-8 \\
weight decay & 0.05 \\

input resolution & 336\suptext{2} / 336\suptext{2} / 448\suptext{2} / 448\suptext{2} \\
patch size & 14\suptext{2} \\
batch size & 1024 \\
training epochs & 100 / 100 / 30 / 30 \\
warmup epochs & 5 / 5 / 3 / 3 \\

drop path~\cite{huang2016deep} & 0.10 / 0.10 / 0.10 / 0.15  \\
label smoothing~\cite{szegedy2016rethinking} & 0.1 / 0.1 / 0.1 / 0.2 \\

augmentation & RandAug (9, 0.5)~\cite{cubuk2020randaugment} \\
random resized crop & (0.08, 1) \\
test crop ratio & 1.0 \\

numerical precision & $\mathtt{DeepSpeed}$ $\mathtt{fp16}$~\cite{rasley2020deepspeed} \\
ZeRO optimizer~\cite{ramesh2021zero} & stage 0 or 1 \\

ema~\cite{ema} & 0.9999 \\
\demphs{cutmix~\cite{yun2019cutmix}} & \demphs{\xmark} \\
\demphs{mixup~\cite{zhang2017mixup}} & \demphs{\xmark} \\
\demphs{random erasing~\cite{zhong2020random}} & \demphs{\xmark} \\

\end{tabular}
\vspace{-.5em}
\caption{Fine-tuning setting for \textbf{IN-1K} (\textbf{\textit{w/o}} \textbf{IN-21K} intermediate fine-tuning).}
\label{tab: 1k ft cfg}
\end{table}

\begin{table}[t!]
\centering
\tablestyle{8pt}{1.2}
\scriptsize
\begin{tabular}{l|c}
config & \eva-B / -L / -L+ \\
\shline

image enc. weight init. & \eva-B / -L / \eva-CLIP-L \\
text enc. weight init. & OpenAI CLIP-B / -L / \eva-CLIP-L \\

image-text data & LAION-1.6B~\cite{laion5b} + COYO-0.4B~\cite{kakaobrain2022coyo-700m} \\

image enc. peak learning rate &  2e-4 / 4e-4 / 4e-4 \\
image enc. layer-wise lr decay~\cite{clark2020electra, bao2021beit} & 0.75 / 0.85 / 0.75 \\
text enc. peak learning rate &  2e-5 / 4e-5 / 4e-5 \\
text enc. layer-wise lr decay~\cite{clark2020electra, bao2021beit} & 0.75 / 0.75 / 0.65 \\

learning rate schedule & cosine decay \\

optimizer & LAMB~\cite{lamb} \\
optimizer hyper-parameters & $\beta_1$, $\beta_2$, $\epsilon$ = 0.9, 0.98, 1e-6 \\
weight decay & 0.05 \\

input resolution & 224\suptext{2} / 224\suptext{2} / 336\suptext{2} \\
patch size & 16\suptext{2} / 14\suptext{2} / 14\suptext{2} \\

batch size & 131k / 131k / 61k \\
samples seen & 8B / 4B / 2B \\

random resized crop & (0.9, 1) \\

numerical precision & $\mathtt{DeepSpeed}$ $\mathtt{fp16}$~\cite{rasley2020deepspeed} \\
ZeRO optimizer~\cite{ramesh2021zero} & stage 1 \\

\demphs{drop path}~\cite{huang2016deep} & \demphs{\xmark} \\
\demphs{FLIP training}~\cite{li2022scaling} & \demphs{\xmark} \\
\demphs{ema~\cite{ema}} & \demphs{\xmark} \\
\demphs{image augmentation} & \demphs{\xmark} \\
\demphs{image cutmix~\cite{yun2019cutmix}} & \demphs{\xmark} \\
\demphs{image mixup~\cite{zhang2017mixup}} & \demphs{\xmark} \\
\demphs{image random erasing~\cite{zhong2020random}} & \demphs{\xmark} \\

\end{tabular}
\vspace{-.5em}
\caption{\eva enhanced Contrastive Language-Image Pre-training (CLIP) setting.}
\label{tab: clip cfg}
\end{table}

\begin{table}[t!]
\centering
\tablestyle{8pt}{1.2}
\scriptsize
\begin{tabular}{l|c}
config & \eva-L \\
\shline

enc. weight initialization & MIM pre-trained \eva (\tblref{tab: pt cfg}) \\

learning rate & 6e-5 \\
layer-wise lr decay & 0.8 \\
batch size & 160 \\
training steps & 400k \\
learning rate schedule & lr step at [320k, 360k] \\

optimizer & AdamW~\cite{adam,Loshchilov2019adamw} \\
optimizer hyper-parameters & $\beta_1$, $\beta_2$, $\epsilon$ = 0.9, 0.999, 1e-8 \\
weight decay & 0.1 \\

LSJ~\cite{simple_copy_paste} crop size & 1536\suptext{2} \\
patch size & 16\suptext{2} \\ 
attention window size & 16\suptext{2} \\
\#global attention blocks & evenly 8 blocks \\

drop path & 0.4 \\

numerical precision & $\mathtt{PyTorch}$ $\mathtt{amp}$ $\mathtt{fp16}$~\cite{pytorch} \\

\demphs{ema}~\cite{ema} & \demphs{\xmark} \\

\end{tabular}
\caption{\textbf{O365} object detection and instance segmentation \textbf{intermediate fine-tuning} setting based on ViTDet~\cite{li2022exploring}.}
\label{tab: o365 cfg}    
\end{table}

\begin{table}[t!]
\centering
\tablestyle{8pt}{1.2}
\scriptsize
\begin{tabular}{l|c}
config & \eva-B / -L \\
\shline

enc. weight initialization & MIM pre-trained \eva (\tblref{tab: pt cfg}) \\

learning rate & 5e-5 / 6e-5 \\
layer-wise lr decay & 0.7 / 0.8 \\
batch size & 128 / 144 \\
training steps & 60k \\
learning rate schedule & lr step at [48k, 54k] \\

optimizer & AdamW~\cite{adam,Loshchilov2019adamw} \\
optimizer hyper-parameters & $\beta_1$, $\beta_2$, $\epsilon$ = 0.9, 0.999, 1e-8 \\
weight decay & 0.1 \\

LSJ~\cite{simple_copy_paste} crop size & 1024\suptext{2} \\
patch size & 16\suptext{2} \\ 
attention window size & 16\suptext{2} \\
\#global attention blocks & evenly 4 blocks \\

drop path & 0.1 / 0.4 \\
test score threshold & 0.05 \\
max numbers of detection & 100 \\

numerical precision & $\mathtt{PyTorch}$ $\mathtt{amp}$ $\mathtt{fp16}$~\cite{pytorch} \\

\demphs{softnms}~\cite{bodla2017soft} & \demphs{\xmark} \\
\demphs{maskness scoring}~\cite{huang2019mask,solo} & \demphs{\xmark} \\
\demphs{ema}~\cite{ema} & \demphs{\xmark} \\

\end{tabular}
\caption{\textbf{COCO} object detection and instance segmentation, \textbf{head-to-head} comparisons setting based on ViTDet~\cite{li2022exploring}.}
\label{tab: coco head to head cfg}
\end{table}

\begin{table}[t!]
\centering
\tablestyle{8pt}{1.2}
\scriptsize
\begin{tabular}{l|c}
config & \eva-B / -L \\
\shline

enc. weight initialization & MIM pre-trained \eva (\tblref{tab: pt cfg}) \\

learning rate & 1e-4 \\
layer-wise lr decay & 0.7 / 0.8 \\
batch size & 128 \\
training steps & 50k / 40k \\
learning rate schedule & lr step at [40k, 45k] / [32k, 36k] \\

optimizer & AdamW~\cite{adam,Loshchilov2019adamw} \\
optimizer hyper-parameters & $\beta_1$, $\beta_2$, $\epsilon$ = 0.9, 0.999, 1e-8 \\
weight decay & 0.1 \\

LSJ~\cite{simple_copy_paste} crop size & 1024\suptext{2} \\
patch size & 16\suptext{2} \\ 
attention window size & 16\suptext{2} \\
\#global attention blocks & evenly 4 blocks \\

drop path & 0.1 / 0.4 \\
test score threshold & 0.02 \\

numerical precision & $\mathtt{PyTorch}$ $\mathtt{amp}$ $\mathtt{fp16}$~\cite{pytorch} \\

\demphs{softnms}~\cite{bodla2017soft} & \demphs{\xmark} \\
\demphs{maskness scoring}~\cite{huang2019mask,solo} & \demphs{\xmark} \\
\demphs{ema}~\cite{ema} & \demphs{\xmark} \\

\end{tabular}
\caption{\textbf{LVIS} object detection and instance segmentation, \textbf{head-to-head} comparisons setting based on ViTDet~\cite{li2022exploring}.}
\label{tab: lvis head to head cfg}
\end{table}

\begin{table}[t!]
\centering
\tablestyle{8pt}{1.2}
\scriptsize
\begin{tabular}{l|c}
config & \eva-B / -L \\
\shline

enc. weight initialization & MIM pre-trained \eva (\tblref{tab: pt cfg}) \\

learning rate & 5e-5 \\
layer-wise lr decay & 0.7 / 0.8 \\
batch size & 128 \\
training steps & 60k \\
learning rate schedule & lr step at [48k, 54k] \\

optimizer & AdamW~\cite{adam,Loshchilov2019adamw} \\
optimizer hyper-parameters & $\beta_1$, $\beta_2$, $\epsilon$ = 0.9, 0.999, 1e-8 \\
weight decay & 0.1 \\

LSJ~\cite{simple_copy_paste} crop size & 1536\suptext{2} \\
patch size & 16\suptext{2} \\ 
attention window size & 32\suptext{2} \\
\#global attention blocks & evenly 6 / 8 blocks \\

drop path & 0.1 / 0.4 \\
test score threshold & 0.00 \\
max numbers of detection & 100 \\

softnms~\cite{bodla2017soft} & IoU threshold = 0.6 \\
maskness scoring~\cite{huang2019mask,solo} & maskness threshold = 0.5 (instance seg only) \\
\demphs{ema}~\cite{ema} & \demphs{\xmark} \\

numerical precision & $\mathtt{PyTorch}$ $\mathtt{amp}$ $\mathtt{fp16}$~\cite{pytorch} \\

\end{tabular}
\caption{\textbf{COCO} object detection and instance segmentation, \textbf{system-level} comparisons setting based on ViTDet~\cite{li2022exploring} (\textbf{\textit{w/o}} O365 intermediate fine-tuning).}
\label{tab: coco wo o365 sota cfg}    
\end{table}

\begin{table}[t!]
\centering
\tablestyle{8pt}{1.2}
\scriptsize
\begin{tabular}{l|c}
config & \eva-L \\
\shline

enc. weight initialization & MIM pre-trained \eva (\tblref{tab: pt cfg}) \\

learning rate & 1e-4 \\
layer-wise lr decay & 0.8 \\
batch size & 128 \\
training steps & 40k \\
learning rate schedule & lr step at [32k, 36k] \\

optimizer & AdamW~\cite{adam,Loshchilov2019adamw} \\
optimizer hyper-parameters & $\beta_1$, $\beta_2$, $\epsilon$ = 0.9, 0.999, 1e-8 \\
weight decay & 0.1 \\

LSJ~\cite{simple_copy_paste} crop size & 1536\suptext{2} \\
patch size & 16\suptext{2} \\ 
attention window size & 32\suptext{2} \\
\#global attention blocks & evenly 8 blocks \\

drop path & 0.4 \\
test score threshold & 0.02 \\
max numbers of detection & 300 \\

softnms~\cite{bodla2017soft} & IoU threshold = 0.6 \\
maskness scoring~\cite{huang2019mask,solo} & maskness threshold = 0.5 \\

numerical precision & $\mathtt{PyTorch}$ $\mathtt{amp}$ $\mathtt{fp16}$~\cite{pytorch} \\

\demphs{ema}~\cite{ema} & \demphs{\xmark} \\

\end{tabular}
\caption{\textbf{LVIS} object detection and instance segmentation, \textbf{system-level} comparisons setting based on ViTDet~\cite{li2022exploring} (\textbf{\textit{w/o}} O365 intermediate fine-tuning).}
\label{tab: lvis wo o365 sota cfg}    
\end{table}

\begin{table}[t!]
\centering
\tablestyle{8pt}{1.2}
\scriptsize
\begin{tabular}{l|c}
config & \eva-L \\
\shline

enc. weight initialization & O365 fine-tuned \eva (\tblref{tab: o365 cfg}) \\

learning rate & 4e-5 \\
layer-wise lr decay & 0.8 \\
batch size & 64 \\
training steps & 40k \\
learning rate schedule & constant \\

optimizer & AdamW~\cite{adam,Loshchilov2019adamw} \\
optimizer hyper-parameters & $\beta_1$, $\beta_2$, $\epsilon$ = 0.9, 0.999, 1e-8 \\
weight decay & 0.1 \\

LSJ~\cite{simple_copy_paste} crop size & 1536\suptext{2} \\
patch size & 16\suptext{2} \\ 
attention window size & 16\suptext{2} \\
\#global attention blocks & evenly 8 blocks \\

drop path & 0.3 \\
test score threshold & 0.00 \\
max numbers of detection & 100 \\

softnms~\cite{bodla2017soft} & IoU threshold = 0.6 \\
maskness scoring~\cite{huang2019mask,solo} & maskness threshold = 0.5 (instance seg only) \\

numerical precision & $\mathtt{PyTorch}$ $\mathtt{amp}$ $\mathtt{fp16}$~\cite{pytorch} \\

ema~\cite{ema} & 0.9999 \\

\end{tabular}
\caption{\textbf{COCO} object detection and instance segmentation, \textbf{system-level} comparisons setting based on ViTDet~\cite{li2022exploring} (\textbf{\textit{w/}} O365 intermediate fine-tuning).}
\label{tab: coco w o365 sota cfg}    
\end{table}

\begin{table}[t!]
\centering
\tablestyle{8pt}{1.2}
\scriptsize
\begin{tabular}{l|c}
config & \eva-L \\
\shline

enc. weight initialization & O365 fine-tuned \eva (\tblref{tab: o365 cfg}) \\

learning rate & 4e-5 \\
layer-wise lr decay & 0.8 \\
batch size & 64 \\
training steps & 70k \\
learning rate schedule & constant \\

optimizer & AdamW~\cite{adam,Loshchilov2019adamw} \\
optimizer hyper-parameters & $\beta_1$, $\beta_2$, $\epsilon$ = 0.9, 0.999, 1e-8 \\
weight decay & 0.1 \\

LSJ~\cite{simple_copy_paste} crop size & 1536\suptext{2} \\
patch size & 16\suptext{2} \\ 
attention window size & 16\suptext{2} \\
\#global attention blocks & evenly 8 blocks \\

drop path & 0.3 \\
test score threshold & 0.02 \\
max numbers of detection & 1000 \\

softnms~\cite{bodla2017soft} & IoU threshold = 0.6 \\
maskness scoring~\cite{huang2019mask,solo} & maskness threshold = 0.5 \\

numerical precision & $\mathtt{PyTorch}$ $\mathtt{amp}$ $\mathtt{fp16}$~\cite{pytorch} \\

ema~\cite{ema} & 0.9999 \\

\end{tabular}
\caption{\textbf{LVIS} object detection and instance segmentation, \textbf{system-level} comparisons setting based on ViTDet~\cite{li2022exploring} (\textbf{\textit{w/}} O365 intermediate fine-tuning).}
\label{tab: lvis w o365 sota cfg}    
\end{table}

\begin{table}[t!]
\centering
\tablestyle{8pt}{1.2}
\scriptsize
\begin{tabular}{l|c}
config & \eva-B / -L / -L+ \\
\shline

enc. weight initialization & MIM pre-trained \eva (\tblref{tab: pt cfg}) \\

learning rate & 6e-5 / 4e-5 / 4e-5 \\
layer-wise lr decay & 0.85 / 0.90 / 0.90 \\
batch size & 32 / 16 / 16 \\
training steps & 60k / 80k / 80k \\
learning rate schedule & linear decay \\

optimizer & AdamW~\cite{adam,Loshchilov2019adamw} \\
optimizer hyper-parameters & $\beta_1$, $\beta_2$, $\epsilon$ = 0.9, 0.999, 1e-8 \\
weight decay & 0.05 \\

crop size & 512\suptext{2} / 512\suptext{2} / 640\suptext{2} \\
patch size & 16\suptext{2} \\ 

drop path & 0.15 / 0.20 / 0.20 \\
seg head dim & 768 / 1024 / 1536 \\

numerical precision & $\mathtt{PyTorch}$ $\mathtt{amp}$ $\mathtt{fp16}$~\cite{pytorch} \\

\demphs{ViT-Adapter~\cite{vitadapt}} & \demphs{\xmark}

\end{tabular}
\caption{Semantic segmentation on \textbf{ADE20K} using \textbf{UperNet}~\cite{xiao2018upernet}.}
\label{tab: seg upernet ade cfg}    
\end{table}

\begin{table}[t!]
\centering
\tablestyle{8pt}{1.2}
\scriptsize
\begin{tabular}{l|c}
config & \eva-L \\
\shline

enc. weight initialization & MIM pre-trained \eva (\tblref{tab: pt cfg}) \\

learning rate & 2e-5 \\
layer-wise lr decay & 0.9 \\
batch size & 16 \\
training steps & 120k \\
learning rate schedule & linear decay \\

optimizer & AdamW~\cite{adam,Loshchilov2019adamw} \\
optimizer hyper-parameters & $\beta_1$, $\beta_2$, $\epsilon$ = 0.9, 0.999, 1e-8 \\
weight decay & 0.05 \\

crop size & 640\suptext{2} \\
patch size & 16\suptext{2} \\ 

drop path & 0.2 \\
seg head dim & 1024 \\
seg head \#enc. \& \#dec. & 6 \& 9 \\

numerical precision & $\mathtt{PyTorch}$ $\mathtt{amp}$ $\mathtt{fp16}$~\cite{pytorch} \\

\demphs{ViT-Adapter~\cite{vitadapt}} & \demphs{\xmark} 

\end{tabular}
\caption{Semantic segmentation on \textbf{COCO-Stuff-164K} using \textbf{Mask2Former}~\cite{mask2former}.}
\label{tab: seg m2f coco cfg}    
\end{table}

\begin{table}[t!]
\centering
\tablestyle{8pt}{1.2}
\scriptsize
\begin{tabular}{l|c}
config & \eva-L \\
\shline

enc. weight initialization & COCO-Stuff fine-tuned \eva (\tblref{tab: seg m2f coco cfg}) \\

learning rate & 2e-5 \\
layer-wise lr decay & 0.9 \\
batch size & 64 \\
training steps & 20k \\
learning rate schedule & linear decay \\

optimizer & AdamW~\cite{adam,Loshchilov2019adamw} \\
optimizer hyper-parameters & $\beta_1$, $\beta_2$, $\epsilon$ = 0.9, 0.999, 1e-8 \\
weight decay & 0.05 \\

crop size & 640\suptext{2} \\
patch size & 16\suptext{2} \\ 

drop path & 0.2 \\
seg head dim & 1024 \\
seg head \#enc. \& \#dec. & 6 \& 9 \\

numerical precision & $\mathtt{PyTorch}$ $\mathtt{amp}$ $\mathtt{fp16}$~\cite{pytorch} \\

\demphs{ViT-Adapter~\cite{vitadapt}} & \demphs{\xmark} 

\end{tabular}
\caption{Semantic segmentation on \textbf{COCO-Stuff-164K} using \textbf{Mask2Former}~\cite{mask2former}.}
\label{tab: seg m2f ade cfg}    
\end{table}

\clearpage

{
\fontsize{8.2pt}{9.84pt}\selectfont
\bibliographystyle{ieee_fullname}
\bibliography{eva02}
}

\end{document}